\documentclass[twocolumn]{svjour3}

\usepackage{francois-preamble,hyperref}
\usepackage{wrapfig}
\usepackage{bm}
\usepackage{algorithmic}
\usepackage{array}

\graphicspath{{figures/}}

\usepackage[caption=false,font=footnotesize,labelfont=sf,textfont=sf]{subfig}

\usepackage{url}

\makeatletter
\def\cl@chapter{\@elt {theorem}}
\makeatother

\usepackage{ifthen} 
\newboolean{Drafty}%
\setboolean{Drafty}{true}%
\ifthenelse{\boolean{Drafty}}%
{%
    \newcommand{\francois}[1]{{\color{red}\textbf{Francois: }#1}}
    \newcommand{\Sune}[1]{{\color{green}\textbf{Sune: }#1}} 
}%
{%
    \newcommand{\francois}[1]{{}}
    \newcommand{\Sune}[1]{{}} 
}%

\newboolean{MarkChanges}%
\setboolean{MarkChanges}{false}%
\ifthenelse{\boolean{MarkChanges}}%
{
	\newcommand{\revchange}[1]{{\color{blue}#1}}%
}%
{%
	\newcommand{\revchange}[1]{{#1}}%
}%

\usepackage{enumitem}
\usepackage{float}
\usepackage{cleveref}
\usepackage{todonotes}
\usepackage{tikz}
\usepackage{tikz-cd}

\author{Henrik G. Jensen \and Fran\c{c}ois Lauze \and Sune Darkner}
\institute{H. G. Jensen \at Dept. of Computer Science, University of Copenhagen \\
  \email{henrikgjensen@gmail.com}
  \and F. Lauze \at
  Dept. of Computer Science, University of Copenhagen \\
  \email{francois@di.ku.dk}           
  \and
  S. Darkner\at
  Dept. of Computer Science, University of Copenhagen \\
  \email{darkner@diku.dk}           
  }

\begin{document}

\title{Information-Theoretic Registration with Explicit Reorientation of Diffusion-Weighted Images}

\maketitle

\begin{abstract}
  %
  We present an information-theoretic appro\-ach to the registration of images with directional information, and especially for diffusion-Weighted Images (DWI), with explicit optimization over the directional scale. We call it Locally Orderless Registration with Directions (LORD). We focus on normalized mutual information as a robust information-theoretic similarity measure for DWI. The framework is an extension of the LOR-DWI density-based hierarchical scale-space model that varies and optimizes the integration, spatial, directional, and intensity scales. As affine transformations are insufficient for inter-subject registration, we extend the model to non-rigid deformations.  We illustrate that the proposed model deforms orientation distribution functions (ODFs) correctly and is capable of handling the classic complex challenges in DWI-registrations, such as the registration of fiber-crossings along with kissing, fanning, and interleaving fibers. \revchange{Our experimental results} clearly illustrate a novel promising regularizing effect, that comes from the nonlinear orientation-based cost function. We show the properties of the different image scales and, we show that including orientational information in our model makes the model better at retrieving deformations in contrast to standard scalar-based registration.
\end{abstract}



%

\section{Introduction}
In this work, we study registration problems for images of the type $I:\Omega\to \Ff(\PP^2)$ with 
$\Omega$ an open, bounded domain of $\RR^3$, the image domain, $\PP^2$ is the projective plane of 
directions (vector lines) in $\RR^3$ and $\Ff(\PP^2)$ is a suitable function space on $\PP^2$: each $I(\bsx)$ is a 
function  $I(\bsx):\PP^2\to\RR$ which provides a measure of a  phenomenon along a three-dimensional 
direction. We may and will also reinterpret such an image $I$ as an image 
$\bI:\Omega\times\PP^2\to\RR$, $\bI(\bsx, \bsv) = I(\bsx)(\bsv)$. 
These functions can be used to represent orientation distribution functions used in Diffusion-Weighted Imaging (DWI).

DWI is a non-invasive Magnetic Resonance Imaging (MRI) protocol that can be used to infer microstructures of biological tissues by tracking the movement of water molecules, otherwise invisible in structural MRI. However, the spatio-directional geometry of DWI signals makes it a challenge in image registration, a key tool for comparing and analyzing medical image data. Also, DWI acquired on different scanners or with different protocols have a non-linear relationship, and defining the similarity between two DWIs is inherently \revchange{difficult \cite{johansen2013diffusion}}. Information-theoretic similarity measures such as mutual information (MI) and normalized mutual information (NMI) \revchange{have} been shown to handle both linear and non-linear relationships very well \cite{darkner2013locally,studholme1999overlap,darkner2018collocation} and are therefore attractive for use in the registration of DWI. 

Our contribution is a scale-space formulation of density estimation for image similarity for DWI in a nonrigid registration setting. The model encompasses explicit reorientation providing the computational framework for estimating non-linear similarity measures well-suited for DWI. The model allows for image similarities such as NMI by extending the Locally Orderless Registration for Diffusion-Weighted Images (LOR-DWI) \cite{jensen2015locally} to nonrigid registration as affine transformations provide insufficient flexibility to perform an accurate inter-subject registration. LORD includes directional information in the image similarity, which allows explicit optimization over the reorientation of diffusion gradients. This extends from diffusion tensor images (DTI) to the raw high-angular resolution diffusion images (HARDI) or the topographically inverted Orientation Density\linebreak Functions (ODF).

To demonstrate \revchange{the capabilities of the framework}, we first build an objective function over a spline  base parametric deformation field, from NMI and a simple regularization, in a multi-resolution setting. We partially compute the gradient of the objective -- as some of the computations are standard -- and indicate how it is optimized.
Then we experimentally study some of its properties, such as the effect of the reorientation on the ODFs, the effect of the scale-space on optimization and regularization on simulated DWI data and on a synthetic deformation of data from a subject of the Human Connectome Project (HCP) \cite{van2013wu}. We show that LORD preserves the ODFs and produces excellent mappings for crossing, kissing and curving fibers, while \revchange{empirically providing an inherent regularizing effect}. A example PyTorch implementation is provided in \url{https://github.com/sunedarkner/Registration}.

The density formulation also allows us to optimize over the isocurves of the DWI signal $I:\mathbb{R}^3\times\mathbb{P}^2\rightarrow\mathbb{R}$.  In principle, this framework could be extended to other non-trivial geometry of other image modalities where scale-space structures can be defined. In this work, we focus on DWI data where we use a classical scale-space structure on this non-linear geometry.

\section{Related Work}
This work addresses two of the major challenges in the voxel-based registration of DWI: the reorientation of DWI in image registration, and the non-linear similarity between DWI.

Image registration refers to a process that transforms data into a shared coordinate system. For DWI, the common way to register two images is to use scalar-based methods on quantitative measures, such as the fractional anisotropy (FA) or the mean diffusivity (MD) \cite{o2017advances}. However, as such methods disregard most of the directional information in DWI, methods have been developed to also account for the reorientation of the diffusion profile. Most of these methods are created on top of scalar-based methods and iteratively reorients the gradients after changing the deformation field. Some of the most popular approaches can be found in \cite{avants2009advanced,tournier2012mrtrix,jenkinson2012fsl,fischl2012freesurfer}. However, registration frameworks have also been designed with an objective function that explicitly optimizes over the reorientation of the gradients, such as \texttt{DTI-TK} \cite{zhang2006deformable}, \texttt{DT-REFineD} \cite{yeo2009dt}, and the more recent \texttt{DR-TAMAS} \cite{irfanoglu2016dr}. These frameworks have been shown to generally outperform scalar-based frameworks for DWI \cite{wang2011dti,zhang2014large,wang2016evaluation,wang2017evaluations}. 


For scalar based approaches to image similarity, registration is straightforward as any popular scalar-based measure can be used, e.g. the \revchange{sum-of-squared difference} (SSD) \cite{yeo2009dt} or mutual information (MI) \cite{wells1996mmv,wang2017evaluations}. Explicit reorientation strategies inherently define the similarity over both position and orientation, and once the full diffusion profile is part of the similarity, these similarity-measures can be formulated in a well-suited way for the non-linear relationship between pairs of DWI. Both \cite{zhang2006deformable} and \cite{yeo2009dt} used variations of SSD in the objective function, while MI was used in \cite{jensen2015locally} through an extension of the LOR framework in \cite{darkner2013locally,darknersporring2011ipmi}. As argued in \cite{jensen2015locally}, the invariant and statistical properties of MI makes it a logical choice for DWI, where multiple factors result in a less functional, more statistical relationship, such as variations in $b$-values, non-monoexponential behavior in biological tissues, and inter-scanner variability \cite{johansen2013diffusion}. MI is often used in the standard registration of complex modalities and, as such, in scalar-based registration of DWI \cite{van2007nonrigid}, and in the pre-processing of DWI, \cite{treiber2016characterization}. MI and normalized MI (NMI) \cite{studholme1999overlap} \revchange{provide} a non-linear statistical measure for DWI. It is nevertheless possible that more functional measures can be well-suited, among others, cross-correlation (CC) and normalized CC (NCC), both of which can be defined from the LOR density-formulation \cite{sporring2011jacobians,darknersporring2012pami}. The density-based DWI comes from the generalized way of estimating image similarity measures based on Locally Orderless Images (LOI) \cite{koenderink1999structure}. The first mention of LOI in the context of image registration was in \cite{hermosillo2002variational} where a variational approach to image registration was presented. The LOR framework \cite{darknersporring2011ipmi,darknersporring2012pami} generalized a range of similarity measures as linear and non-linear functions of density estimates for scalar-valued images.

\section{Locally Orderless  Directional Imaging}
\label{sec:lor_dwi}
\subsection{Notations}
$\Omega\subset \mathbb R^3$ is the spatial domain of the images under consideration. A scalar image is a function $\bm I:\Omega\to\mathbb R^3$. We assume that we can extend it on the whole $\mathbb R^3$, for instance by extending it with 0. The convolution of two functions $f, g:\RR^n\to\RR$ is defined as $f*g(\bsx) = \int_{\RR^n}f(\bsy)g(\bsx-\bsy)\,d\bsy$.  The projective space of directions of $\mathbb R^3$, i..e, the set of vector lines of $\RR^3$, is denoted by $\mathbb P^2$, and the unit sphere of $\mathbb R^3$ by $\mathbb S^2$.  We will encounter \emph{spatio-directional} images $\bm I:\Omega\times \mathbb P^2 \to \mathbb R$, which we similarly assume to be extendable to $\mathbb R^3\times \mathbb P^2$. This is necessary in both cases so as to define their spatial smoothing via convolution. We use the following elementary property: As $\mathbb P^2$ is naturally identified as the quotient $\mathbb S^2/\{\pm 1\}$ by the antipodal symmetry, a function $f:\mathbb P^2\to \mathbb R$ can be lifted to an antipodal symmetric function $\tilde{f}:\mathbb{S}^2\to \mathbb{R}$. Conversely, any antipodal symmetric function $g:\mathbb S^2\to \mathbb R$ factors through $\mathbb P^2$. A spatio-directional image can and will be lifted to an antipodal symmetric image $\tilde{\bm I}:\Omega\times \mathbb S^2$: $\tilde{\bm I}(\bsx, -\bsv) =\tilde{\bm I}(\bsx, \bsv)$.  Such a lift allows us to represent $\PP^2$-functions as $\SS^2$ ones, with the benefit that $\SS^2$ is an oriented manifold, while $\PP^2$ is not. This means in particular that $\PP^2$ does not have a global volume form. On the other hand, $\SS^2$ has, and we use the one induced by its standard Riemannian metric.

We will denote by $\bm I$ both the spatio-directional image and its antipodal symmetric lifting in the following.

\subsection{\revchange{Recalling the LOR Framework}}
The LOR framework defines the density estimates over three scales: The image scale $\sigma$, the intensity scale $\beta$, and the integration scale $\alpha$. In the context of scalar registration, for a transformation, $\phi:\mathbb{R}^3\rightarrow\mathbb{R}^3$, the estimated histogram $h$ and the corresponding density $p$ is computed as 
\begin{align}
\label{eq:loih}
  &h_{\beta\alpha\sigma}(i,j|\phi,\bsx)=\nonumber\\
                 &\int_{\RR^3}\!P_\beta( \bm I_{\sigma}(\phi(\bsy))-i)
                   P_\beta( \bm J_{\sigma}(\bsy)-j)W_\alpha(\bsy-\bsx)d\bsy \\
  \label{eq:loipdf}
  &p_{\beta\alpha\sigma}(i,j|\phi,\bsx)\simeq \frac{h_{\beta\alpha\sigma}(i,j|\phi,\bsx)}
                         {\int_{\Lambda^2} h_{\beta\alpha\sigma}(k,l|\phi,\bsx)dk\; dl}
\end{align} 
where $i,j\in[a_1, a_2] =: \Lambda$ are values in the image intensity range, $\bm I_{\sigma}(\bm \phi(\bsx))=(\bm I*K_\sigma)(\bm \phi(\bsx))$ and $ J_{\sigma}(\bsx)=(\bm J*K_\sigma)(\bsx)$ are images convolved with the kernel $K_\sigma$ with standard deviation $\sigma$, $P_\beta$ is a Parzen-window  of scale $\beta$ (\cite{bishop:2006:PRML},  paragraph 2.5.1), and $W_\alpha$ is a Gaussian integration window of scale $\alpha$. The marginals are trivially obtained by integration over the appropriate variable.  The LOR-approach to similarity lets us use a set of generalized similarity measures, linear or non-linear
\begin{align}\label{eq:sim}
  \mathcal{F}_{lin}=\int_{\Lambda^2}f(i,j)p(i,j)di\; dj\\ \mathcal{F}_{non-lin}=\int_{\Lambda^2}f(p(i,j))di\; dj.
\end{align}
The linear measure $f(i,j)$ includes e.g. sum of squared differences and Huber. The non-linear $f(p(i,j))$ includes e.g. MI, normalized MI (NMI), see \cite{darknersporring2012pami} for details.

\subsection{The LORD framework}
This work addresses the estimation of the image similarity $\mathcal F$ of DWI in the context of nonrigid registration an extension of our previous work \cite{jensen2015locally}. In this context, $\bI$ and $\bJ$ are spatio-directional signals. Specifically, DWI MR attenuation signals at location $\bsx$, for a gradient direction $\bsv$, are modeled by $S(\bsx,\bsv) = S_0(\bsx) e^{-b\bm I(\bsx,\bsv)}$ \ \cite{tao2006method} and apparent diffusion coefficients volumes are given by $\bm I(\bsx,\bsv)=-\frac{1}{b}\log\frac{S(\bsx,\bsv)}{S_0(\bsx)}$. Gradient directions $\bsv$ belong to $\mathbb S^2$ but the diffusion is orientation-free, $\bm I(\bsx,\bsv)\approx \bm I(\bsx,-\bsv)$. This  defines naturally a spatio-directional image $\Omega\times \PP^2 \to \RR$.  In order to apply LORD, the histogram and density estimates \Cref{eq:loih} and \Cref{eq:loipdf} must be extended to spatio-directional data, and the action of the spatial transformation on the directions must be defined.

We introduce a kernel on the sphere as an extension to the density estimates of LOR to include directional information. This kernel accounts for directional smoothing and defines the LORD framework: we extend spatial smoothing to be spatio-directional. The directional smoothing preserves the antipodal symmetry, equivalently the projective structure, via a symmetric kernel $\Gamma_\kappa(\bm \nu,\bsv)$ on $\mathbb{S}^2$. We define the smoothed signal $\bm I_{\sigma\kappa}$ at scales $(\sigma,\kappa)$ by
\begin{align}
  \bm I_{\sigma\kappa}(\bsx,\bsv)&=\label{eq:vecfieldsmooth}
      \int_{S^2} \!\!\left( \int_{\mathbb R^3} \bm I(\bsy,\bm \nu)K_\sigma(\bsx-\bsy)d\bsy\right) 
     \Gamma_{\kappa}(\bm \nu, \bsv) d\bm \nu.
\end{align}
Here $K_\sigma$ is a Gaussian kernel with standard deviation $\sigma$.  The integral over $\SS^2$ is sometimes referred to as a \emph{spherical convolution} (though actually a \emph{pseudo-convolution} \cite{GallierQuaintance:2020b}). $\Gamma_{\kappa}(\bm \nu,\bsv)$ is a Watson distribution \cite{jupp1989unified,sra_karp:2013}, on $\mathbb S^2$
\begin{align}
&\Gamma_\kappa(\bm \nu,\bsv)=Ce^{\kappa({\bm \nu}^\top \bsv)^2}\\ &C=M\left(\frac{1}{2},\frac{1}{d},\kappa\right)
\end{align}
where $M(\frac12,\frac{d}2,\kappa)$ the confluent hypergeometric function also called the Kummer function ($d=3$ in our case) \cite{AbramowitzStegun1974}, $\pm\bsv$ the center of the distribution, and $\kappa$ the concentration parameter, which is roughly inverse proportional to the variance on the sphere. Because of the symmetry property of the Watson distribution and the antipodal symmetry of $\bm I(\bsx,\bsv)$, it is clear that $\bm I_{\sigma\kappa}(\bsx,\bsv)$ is antipodal symmetric too. As one alternative, a symmetrized von Mises-Fisher \cite{jupp1989unified} distribution or a symmetrized heat kernel could be considered.

\subsubsection{Action on Orientation}
A diffeomorphic transformation $\phi:\Omega\to \Omega$ acts on directions via its Jacobian: at $\bsx\in \Omega$, $J_{\bsx}\phi$ sends the vector line $\RR\bsv$, $\bsv\not=0$,  to the vector line $\RR J_{\bsx}\phi(\bsv)$, this is  well-defined as $\det(J_{\bsx}\phi)\not = 0$. Thus, for each $\bsx\in \Omega$,  $J_{\bsx}\phi$ gives rise to projective linear transformation $\PP J_{\bsx}\phi$. Using the representation $\PP^2\simeq \SS^2/\{\pm1\}$, we can write $\PP J_{\bsx}\phi:\{\pm \bsv\} \mapsto \{\pm\frac{J_{\bsx}\phi(\bsv)}{|J_{\bsx}\phi(\bsv)|}\}$. We denote by $\psi$ the mapping $(\bsx,\bsv)\mapsto \frac{J_{\bsx}\phi(\bsv)}{|J_{\bsx}\phi(\bsv)|}$. Because it will only appear inside an antipodal symmetric kernel, $\psi_{\bsx} = \psi(\bsx,\cdot)$ represents $\PP{J_{\bsx}\phi}$ \emph{without ambiguity}.

\subsubsection{Density, orientation and transformation}
Because of spatial convolutions, we assume that \revchange{the spatial domain $\Omega$ of $\phi$} can be extended to $\RR^3$, by taking it to be the identity out of a compact set $D$ containing $\Omega$. We set 
\begin{equation}
\Phi:\RR^3\times\SS^2\to \RR^3\times\SS^2,\quad \Phi(\bsx,\bsv) = \left(\phi(\bsx),\psi_{\bsx}(\bsv)\right).
\end{equation}	
In the sequel, we in general omit $\bsx$ in $\psi_{\bsx}$, as the spatial location should be unambiguous. We write the joint histogram and density for similarity in image registration as
\begin{multline}
h_{\bsI\circ\Phi,\bsJ;\sigma\kappa\beta\alpha}(i,j|\bsx)=\\ 
                      \int_{\RR^3 \times \SS^2}P_\beta(\bsI_{\sigma\kappa}(\phi(\bsy),\psi(\bsv))-i)
                        P_\beta(\bsJ_{\sigma\kappa}(\bsy,\bsv)-j) \times\\
                      W_\alpha(\bsy-\bsx)d\bsy d\bsv \label{eq:densityreg}
\end{multline}
\vspace{-5mm}
\begin{multline}
 p_{\bsI\circ\Phi,\bsJ;\sigma\kappa\beta\alpha}(i,j|\bsx)=\frac{h_{\bsI\circ\Phi,\bsJ;\sigma\kappa\beta\alpha}(i,j|\bsx)}
                       {\int_{\Lambda^{2}} h_{\bsI\circ\Phi,\bsJ;\sigma\kappa\beta\alpha}(l,k|\bsx)dl\;dk}\;.
\end{multline}
If one assumes that $W_\alpha$ is a Gaussian kernel with standard deviation $\alpha$ and we let $\alpha\rightarrow\infty$, we obtain full spatial integration, and we can write the joint histogram and joint probability density functions (PDF)
\begin{multline}
\label{eq:histo_fullspatial}
	h_{\bsI\circ\Phi,\bsJ;\sigma\kappa\beta}(i, j)=\\
	\int_{\RR^3\times\SS^2}\!\!\!P_\beta(\bsI_{\sigma\kappa}(\phi(\bsx),\psi(\bsv))-i)
	P_\beta(\bsJ_{\sigma\kappa}(\bsx,\bsv)-j)\,d\bsx d\bsv
\end{multline}
\vspace{-5mm}
\begin{equation}
\label{eq:jointdensity}
	p_{\bsI\circ\Phi,\bsJ;\sigma\kappa\beta}(i,j) = \frac{h_{\bsI\circ\Phi,\bsJ;\sigma\kappa\beta}(i, j)}
	{\int_{\Lambda^2}h_{\bsI\circ\Phi,\bsJ;\sigma\kappa\beta}(l,k)dl\;dk}\;.~~~~~~~\hfil
\end{equation}
Similar formulas are obtained for single histograms\linebreak $h_{\bsK;\sigma\kappa\beta}$ and corresponding  PDF $p_{\bsK;\sigma\kappa\beta}$ with $K = \bsI\circ\Phi$ and $\bsJ$,  they are just the appropriate marginals of the joint distribution. This is the situation we consider in rest of this work.

\subsection{LORD and Free Form Deformations}

\subsubsection{Free-form deformations.} In \cite{jensen2015locally}, we assumed that $\phi$ was a global affine transformation, with \revchange{$\psi_{\bsx} \equiv \psi$ independent} of $\bsx$, being the projectivization of the linear part of $\phi$. In the present work, we assume $\phi$ to be a non-rigid transformation, and we use Rueckert \textit{et al.\/} framework \cite{rueckert1999nonrigid}: the transformation $\phi$ is given as a hierarchical spline representation \emph{affinely} parameterized by a spatial grid ${\bsc}$ of control points.  We follow \cite{rueckert1999nonrigid} and describe briefly the construction and refer the reader to it and references therein for details. Our domain $\Omega$ is chosen to be a box $[0,N_x]\times[0,N_y]\times[0,N_z]$. At a given \emph{spatial resolution $r$}, a 3D-grid  $\bsc^r = (\bsc^r_{ijk})$ of size $(n^r_x,n^r_y,n^r_z)$ of equally spaced (to start with) control points, with spacing distance $\delta_r$, is chosen. This defines a displacement function
\begin{equation}
\label{eq.bslinedepl}
	B^r_{\bsc^r}(\bsx)= \sum_{l,m,n=0}^3 B_l(u)B_m(v)B_n(w)\bsc^r_{i+l,j+m,k+l} 
\end{equation}
with $\bsx = (x,y,z)$, $i = \lfloor x/n^r_x\rfloor$-1, $j = \lfloor y/n^r_y\rfloor-1$, $k = \lfloor z/n^r_z\rfloor -1$, $u = x/n^r_x -i + 1$, $v = y/n^r_y -j + 1$, $w = z/n^r_z -k + 1$ and $B_l, B_m$ and $B_n$ are 1D cubic basis B-spline functions. We build $R$ grids of control points $\bsc = (\bsc^1,\dots,\bsc^R)$  from coarse to fine resolution, with  corresponding $B^r_{\bsc^r}$. We set
\begin{equation}
	\phi_{\bsc}(\bsx) = \bsx + \sum_{r=1}^R B_{\bsc^r}^r(\bsx).
\end{equation} 
All the $B_{\bsc^r}^r$ depend linearly on $\bsc^r$, $\phi_{\bsc}$ can be written as $ \bsx + \bsB(\bsx)\text{vec}(\bsc)$, $\text{vec}(\bsc)$ is a suitable vectorization of the control points, $\text{vec}(\bsc)$ has dimensions $3k$, with $k$ the total number of control points,  and $\bsB(\bsx)\in \RR^{3\times 3k}$. In the sequel, we will use $\bsc$ to denote both the grid of control points and its vectorization $\text{vec}(\bsc)$, so as to avoid too cumbersome notations.

The spatial Jacobian $J_{\bsx}\phi_{\bsc}(\bsv) = \bsv + \left(J_{\bsx}\bsB\cdot \bsv\right)\bsc$. $J_{\bsx}\bsB$ is a \emph{tensor} of size $3\times 3k\times 3$, which by \eqref{eq.bslinedepl} is composed of quadratic and cubic B-spline basis functions. The '$\cdot$' operation is the contraction $\RR^{3\times 3k\times 3}\times \RR^3$, $(r_{uvw},s_w)\mapsto \sum_w r_{uvw}s_w$. We then have 
\begin{equation}
\psi_{\bsc}(\bsv) = \frac{\bsv + \left(J_{\bsx}\bsB\cdot \bsv\right)\bsc}{|\bsv + \left(J_{\bsx}\bsB\cdot \bsv\right)\bsc|}
\end{equation}
and set $\Phi_{\bsc} = (\phi_{\bsc}, \psi_{\bsc})$.

\subsubsection{Registration objective.}
We use the normalized mutual information  (NMI) as a similarity measure
\begin{equation}
\Ff(\bsc) = \text{NMI}(\bsI\circ\Phi_{\bsc}, \bsJ; \sigma\kappa\beta) = \frac{H_{\sigma\kappa\beta}(\bsI\circ\Phi_{\bsc}) + H_{\sigma\kappa\beta}(\bsJ) }{H_{\sigma\beta\kappa}(\bsI\circ\Phi_{\bsc}, \bsJ)}.
\end{equation}The different terms are differential entropies of the joint density \eqref{eq:jointdensity} and its marginals.

We add a regularizer in the form of a simple penalization on the non-uniformity of the control point grid by the squared difference between a point $\bsc_i$ and its direct neighbors. Recall that control points are organized as a family of $R$ grids, from coarse to fine resolution. The regularizer $\mathcal S(\bsc)$ is the sum $\sum_{r = 1}^R \mathcal{S}^r(\bsc^r)$ at each resolution, where $\bsc^r = (c^r_1,\dots c^r_{p_r})^T$ is the grid of control points at resolution level $r$. We denote by $N^r(i)$ the set of indices $j$ such that control point $c^r_j$ is neighbor to control point $c^r_i$ and by $|N^r(i)|$ its cardinal. We set
\begin{equation}
\label{eq:regularizer}
\mathcal{S}^r(\bsc)= -\frac{\lambda_r}{2}\sum_{i=1}^{p_r} \|c^r_i - \frac{1}{|N^r(i)|}\sum_{j\in N^r(i)} c^r_j\|^2\;\; 
\end{equation}
$\lambda_r$ is a strictly positive parameter controlling the degree of smoothness. The negative sign in \Cref{eq:regularizer} comes from the fact that we \emph{maximize} the regularized NMI. We seek the transformation $\Phi_{\bsc^*}$ which maximizes the regularized NMI
\begin{equation}
  \bsc^* = \Argmax_{\bsc} \Mm(\bsc)  =  \Argmax_{\bsc}\left(\Ff(\bsc)+\mathcal S(\bsc)\right).
    \label{eq:argmax}
\end{equation}
We use a quasi-Newton method to compute an optimum of objective function \eqref{eq:argmax}, L-BFGS from \cite{schmidt2005minfunc}.
We adapt the method to our multiresolution setting. At step $r\in\{1\dots,R\}$, all grid points except $c^r$ are kept fixed and we optimize our objective with respect to $c^r$. 

This method requires  the gradient $\nabla\Mm(\bsc^r)$ with respect to the control points vector $\bsc^r$ for each resolution. 
The dependency of $\Mm$ in $\bsc^r$ appears, in the similarity term,  deep inside its definition, this is reflected in its differential. In the next paragraph, we provide some details of its computation for the similarity term, \revchange{and in  \Cref{sec:reg}}, we derive the gradient of the regularization term.

\subsection{Differential of the similarity term} 
\begin{figure}[ht]
\begin{equation*}
\begin{tikzcd}[column sep=0.4cm,row sep=0.5cm]
  ~& NMI( \bsI\circ\Phi_{\bsc},\bsJ;\sigma\kappa\beta)&~\\
  H_{\sigma\kappa\beta}(\bsI\circ\Phi_{\bsc}) \ar[ur]&  
  H_{\sigma\kappa\beta}(\bsJ)\ar[u] & 
  H_{\sigma\kappa\beta}(\bsI\circ\Phi_{\bsc}, \bsJ)\ar[lu]\\
  p_{\bm I\circ\Phi_{\bsc};\sigma\kappa\beta}\ar[u]  & 
  p_{\bm J;\sigma\kappa\beta}\ar[u]  & 
  p_{\bm I\circ\Phi_{\bsc}, \bm J;\sigma\kappa\beta}\ar[u]\\
  ~ & h_{\bm I\circ\Phi_{\bsc},\bm J;\sigma\kappa\beta}\ar[ul]\ar[u]\ar[ur]& ~\\
  \bm I_{\sigma\kappa}(\phi_{\bsc},\psi_{\bsc})\ar[ur]&\psi_{\bsc}\ar[l]&\bm J_{\sigma\kappa}\ar[ul]\\
  \boxed{\bsc}\ar[r ]& \phi_{\bsc}\ar[lu] \ar[u]&
  \ar[from=1-1,to=1-3,color=red,dash,line width=0.65pt,start
  anchor={[xshift=-2.9em,yshift=2ex]center},end
  anchor={[xshift=+3.6em,yshift=2ex]center}]
  \ar[from=1-1,to=4-1,color=red,dash,line width=0.65pt,start
  anchor={[xshift=-2.9em,yshift=2ex]center},end
  anchor={[xshift=-2.9em,yshift=-1.5ex]center}]
  \ar[from=4-1,to=4-3,color=red,dash,line width=0.65pt,start
  anchor={[xshift=-2.9em,yshift=-1.5ex]center},end
  anchor={[xshift=+3.6em,yshift=-1.5ex]center}]
  \ar[from=1-3,to=4-3,color=red,dash,line width=0.65pt,start
  anchor={[xshift=+3.6em,yshift=2ex]center},end
  anchor={[xshift=+3.6em,yshift=-1.5ex]center}]
  \ar[from=5-1,to=5-2,color=blue,dash,line width=0.65pt,start
  anchor={[xshift=-2.6em,yshift=1.7ex]center},end
  anchor={[xshift=+0.9em,yshift=1.7ex]center}]
  \ar[from=6-1,to=6-2,color=blue,dash,line width=0.65pt,start
  anchor={[xshift=-2.6em,yshift=-1.7ex]center},end
  anchor={[xshift=+0.9em,yshift=-1.7ex]center}]
  \ar[from=5-1,to=6-1,color=blue,dash,line width=0.65pt,start
  anchor={[xshift=-2.6em,yshift=1.7ex]center},end
  anchor={[xshift=-2.6em,yshift=-1.7ex]center}]
  \ar[from=5-2,to=6-2,color=blue,dash,line width=0.65pt,start
  anchor={[xshift=+0.9em,yshift=1.7ex]center},end
  anchor={[xshift=+0.9em,yshift=-1.7ex]center}]
\end{tikzcd}
\end{equation*}
\caption{Dependency graph of the nonrigid DWI registration between the moving image $\bm I$ and the target image $\bm J$, with NMI as the similarity measure. The deformation is parameterized by $\bm c$ so that any change in $\bm c$ will eventually affect the total similarity between the two images. The red frame contains MI registration elements that are classical, while the blue frame contains the part which extends the LOR framework to the LORD.}
    \label{fig:dependcygraph}
\end{figure}
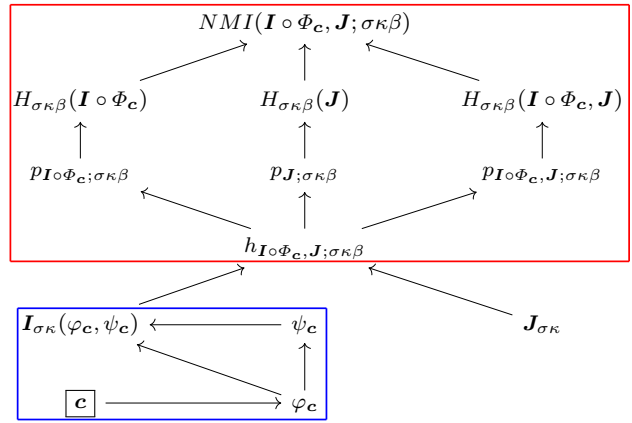

In this paragraph , we omit the subscript $r$ and write $\bsc$ instead of $\bsc^r$. Computations are of the same nature across resolutions. 
The following diagram (\Cref{fig:dependcygraph}) exhibits the different PDFs that are used in the NMI term and illustrates its dependency in $\bsc$, which occurs only through $\Phi_{\bsc}$. Writing $NMI(\bsI\circ\Phi_{\bsc},\bsJ)$ as $\Ll(\bsI_{\sigma\kappa}\circ\Phi_{\bsc})$, its $\bsc$-differential can be written as 
\begin{equation}
	J_{\bsc} NMI(\bsI\circ\Phi_{\bsc}, J) = d_{\bsI}\Ll(\bsI_{\sigma\kappa}\circ\Phi_{\bsc})J_{\bsc}\left(\bsI_{\sigma\kappa}\circ\Phi_{\bsc}\right)
\end{equation}
(we use the notation $d_{\bsI}$ as there is in general no coordinate system on image spaces). A full formula for the Jacobian $J_{\bsc} NMI(\bsI\circ\Phi_{\bsc}, \bsJ)$ (and thus gradient, as its transpose) is long and not very informative. Computations for $d_{\bsI}\Ll$,   correspond to the red frame in  \Cref{fig:dependcygraph}, and are obtained by application of standard derivation rules. They can actually be found in  \cite{rueckert1999nonrigid,darknersporring2012pami,jensen2015locally}. The computations in this paper which differ from the aforementioned correspond to the blue frame in \Cref{fig:dependcygraph} and are now provided.

In our implementation, we replaced the continuous Watson kernel smoothing by a sum over a discrete set of $N$ directions $\nu_1,\dots,\nu_N$ at each voxel $\bsx$. The normalization factor $M(\frac12,\frac13,\kappa)$ in the Watson kernel is replaced by its approximation from the discrete directions and is no longer a constant. \revchange{The spatio-directional smoothing \eqref{eq:vecfieldsmooth} is replaced by the semi-discrete smoothing}
\begin{align}
    \bsI_{\sigma \kappa}&(\phi_{\bsc}(\bsx), \psi_{\bsc}(\bsv)) =\nonumber\\
    &\sum_{n=1}^N \int_{\RR^3}\bsI(\bsy, \nu_n)K_\sigma(\phi_{\bsc}(\bsx)-\bsy)
    \bar{\Gamma}_{\kappa}(\nu_n, \psi_{\bsc}(\bsv))\,d\bsy
\end{align}
where we have set
\begin{equation}
 \label{eq:discreteWatson}
 \bar{\Gamma}_{\kappa}(\nu_n,\psi_{\bsc}(\bsv)) =  \frac{e^{\kappa(\nu_n^\top\psi_{\bsc}(\bsv))^2}}
 {\sum_{i=1}^Ne^{\kappa(\nu_i^\top\psi_{\bsc}(\bsv))^2}} \;\; .
\end{equation}
The Jacobian of the spatio-directional smoo\-thing with respect to the control point parameter $\bm c$ is:
\begin{multline}
    J_{\bsc}\bsI_{\sigma \kappa}(\phi_{\bsc}(\bsx), \psi_{\bsc}(\bsv)) =\\
    \sum_{n=1}^N\left(\int_{\mathbb R^3} \bm I(\bsy, \nu_n)J_{\bsc}K_{\sigma}
    (\phi_{\bsc}(\bsx)-\bsy)\,d\bsy\right)\!\bar{\Gamma}_{\kappa}(\nu_n,\psi_{\bsc}(\bsv))  + \\
    \sum_{n=1}^N\left(\int_{\mathbb R^3} \bm I(\bsy, \nu_n)K_{\sigma}
    (\phi_{\bsc}(\bsx)-\bsy)\,d\bsy\right)\!J_{\bsc}\bar{\Gamma}_{\kappa}(\nu_n,\psi_{\bsc}(\bsv)) 
    \label{eq:sddirs}
\end{multline}
Because $\phi_{\bsc}(\bsx) =\bsx+{\bsB(\bsx)}\bsc$, its Jacobian $J_{\bsc}\phi_{\bsc}(\bsx)$ is simply $\bsB(\bsx)$ and, as $K_\sigma$ is a Gaussian kernel,
 \begin{align}
   J_{\bsc}K_\sigma(\phi_{\bsc}(\bsx)\!-\!\bsy) = -\frac{K_\sigma(\bm\phi_{\bsc}(\bsx)\!-\!\bsy)}{\sigma^2}
   \left(\phi_{\bsc}(\bsx)-\bsy\right)^\top\!\!\!{\bsB(\bsx)}
 \end{align}
 The part involving directional kernel is somewhat more complex, as it involves $\psi_{\bsc}$ and thus the Jacobian $J_{\bsx}\phi_{\bsc}$, as well as the normalizing factor in \Cref{eq:discreteWatson}. Set $f(x) = e^{\kappa x^2}$ and define $f_i = f(\nu_i^\top \psi_{\bsc}(\bsv))$. Since $f'(x) = 2\kappa x f(x)$, we have
\begin{equation}
  J_{\bsc} f(\nu^\top\psi_{\bsc}(\bsv)) = 2\kappa
  f(\nu^\top\psi_{\bsc}(\bsv))\psi_{\bsc}(\bsv)^\top \nu\nu^\top J_{\bsc}\psi_{\bsc}(\bsv) \;\; .
\end{equation}
Then, by a straightforward calculation, we obtain
\begin{align}
    ~&J_{\bsc}\bar{\Gamma}_{\kappa}(\nu_n,\psi_{\bsc}(\bsv))=\nonumber\\
    &-2\kappa\frac{\bar{\Gamma}_{\kappa}(\nu_n,\psi_{\bsc}(\bsv))}{\sum_{i=1}^N f_i}
    \psi_{\bsc}(\bsv)^\top\!\!\left(\sum_{i\not=n}\! f_i\nu_i\nu_i^\top\!\right)\!J_{\bsc}\psi_{\bsc}(\bsv).
\end{align}
The vector $\psi_{\bsc}(\bsv)$ is the normalization of $\bsV := J_{\bsx}\phi_{\bsc}(\bsv)$ $= \bsv + \left(J_{\bsx}\bsB\cdot\bsv\right)\bsc$. The differential of the normalization  $\bsV\mapsto \frac{\bsV}{|\bsV|}$ is $\frac{1}{|\bsV|}\bspi_{\bsV^\bot}$ with $\bspi_{\bsV^\bot}$ the orthogonal projection on $\bsV^\bot$, the subspace orthogonal to $\bsV$.   The Jacobian of $J_{\bsc}J_{\bsx}\phi_{\bsc}(\bsv)$ is simply $J_{\bsx}\bsB\cdot\bsv$ and
\begin{equation}J_{\bsc}\psi_{\bsc}(\bsv) = \frac{1}{|\bsV|}\pi_{\bsV^\bot}J_{\bsx}\bsB\cdot\bsv
\end{equation}

Putting things together, the Jacobian with respect to  the control point parameter $\bm c$ of the spatio-directional smoothing is given by
\begin{multline}
\label{eq:gradF}
   \! \!\!~J_{\bsc}\bsI_{\sigma \kappa}(\phi_{\bsc}(\bsx), \psi_{\bsc}(\bsv)) =\\
    \!\!-\sum_{n=1}^N\int_{\RR^3} \bsI(\bsy, \nu_n)K_\sigma(\phi_{c}(\bsx)-\bsy)\bar{\Gamma}_\kappa(\nu_n,\psi_{\bsc}(\bsv))\times\\
   \left\{\frac{(\phi_{\bsc}(\bsx) - \bsy)^\top\bsB(\bsx)}{\sigma^2}+\right.\\
   \left.\frac{2\kappa \bsV^\top\!\!\left(\!\sum_{i\not=n}f_i\nu_i\nu_i^\top\!\right)}{|\bsV|\sum_{i=1}^N f_i}
  {\bspi}\!_{\bsV\!^\bot}\!\!\left(\!\!\frac{J_{\bsx}\bsB\bullet \bsv}{|\bsV|}\!\right)  
  \!\! \right\}\!d\bsy.
\end{multline}
The gradient of the similarity term is
\begin{equation}
\label{eq:gradF2}
	\nabla\Ff(\bsc)= \left(d_{\bsI}\Ll(\bsI_{\sigma\kappa}\circ\Phi_{\bsc})J_{\bsc}\bsI_{\sigma \kappa}\circ\Phi_{\bsc}\right)^T. 
\end{equation}
The operator $d_{\bsI}\Ll(\bsI_{\sigma\kappa}\circ\Phi_{\bsc})$ acts among other things by spatial, directional and intensity integration of the local changes described in $J_{\bsc}\left(\bsI_{\sigma\kappa}\circ\Phi_{\bsc}\right)$. It is rather complex and makes it difficult to provide an intuitive explanation for the different terms which compose \Cref{eq:gradF}.

\subsection{Regularization}
\label{sec:reg}
In order to compute the gradient of  $\mathcal{S}^r(\bsc^r)$ defined in \Cref{eq:regularizer}, we introduce linear mappings $T^r_i:\bm c\mapsto |N^r(i)|c^r_i - \sum_{j\in N^r(i)}c^r_j$. The regularizer in \Cref{eq:regularizer} can be rewritten as
\begin{equation}
\mathcal{S}^r(\bsc^r)=-\frac{\lambda_r}{2}\sum_{i=1}^{p_r}\frac{1}{|N^r(i)|^2}\|T^r_i\bm c^r\|^2
\end{equation}
and by classical manipulation, we obtain that
\begin{equation}
    \label{eq:reggradient}
    \nabla\mathcal{S}^r(\bsc^r) = -\lambda_r\sum_{i=1}^{p_r}\frac{1}{|N^r(i)|^2}T_i^{r*} T_i^r\bm c^r \;\; .
\end{equation}
$T^{r*}_i$ is the adjoint of $T^r_i$, and given by
$(T_i^* \bsd)_j = N^r(i)\bsd$ if $j=i$, $-\bsd$ if $j\in N^r(i)$ and $0$ otherwise.
The resulting operator $\Delta_r=-\sum_{i=1}^{p_r}\frac{1}{|N^r(i)|^2}T_i^{r*} T^r_i$ is a discrete Laplacian. The sought gradient is
\begin{equation}
    \nabla\mathcal{S}(\bsc) = -\sum_{r = 1}^R\lambda_r\sum_{i=1}^{p_r}\Delta_r\bsc^r.
\end{equation}
In this work, we chose $\lambda_1=\dots=\lambda_R$.
\subsubsection{Implementation details}
To ensure that the deformation is a diffeomorphism, we check the Jacobian of the deformation for negative eigenvalues. In case of a \revchange{negative eigenvalue} of the Jacobian we perform backtracking, which in practice is performed by L-BFGS internally.

\section{Experiments}

To illustrate the properties of the LORD method, we conduct a series of experiments on simulated data and artificially warped real data.

\subsection{Simulated Examples}
\label{sec:artitest}

Artificially generated samples enable visual inspection \revchange{and validation of the} framework in a highly controlled environment. While artificial experiments can be found in most DWI registration papers, artificial data is rarely available. To our knowledge, there are no open sources of simulated DWI data for comparing registration frameworks, although the \texttt{DIPY} project may be a good source for generating simulations \cite{garyfallidis2014dipy}. To this end, we created our own simulated DWI data, which is freely available. The generation of the artificial data is described next.\footnote{Contact us or \url{henrikgjensen@gmail.com} for the code or examples.} 

\subsubsection{Simulating DWI data}
\begin{figure}[ht]
  \centering \subfloat[HARDI
  sample]{\label{subfig:disc}\includegraphics[width=0.3\columnwidth]{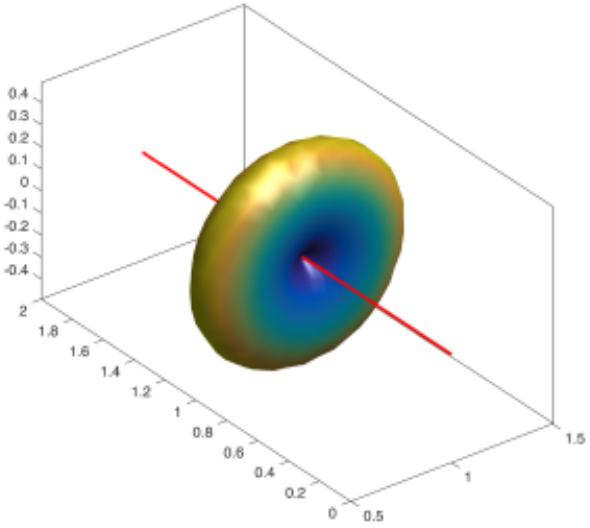}}
  \hspace*{0.7cm}
  \subfloat[ODF of (a)]{\label{subfig:line}\includegraphics[width=0.3\columnwidth]{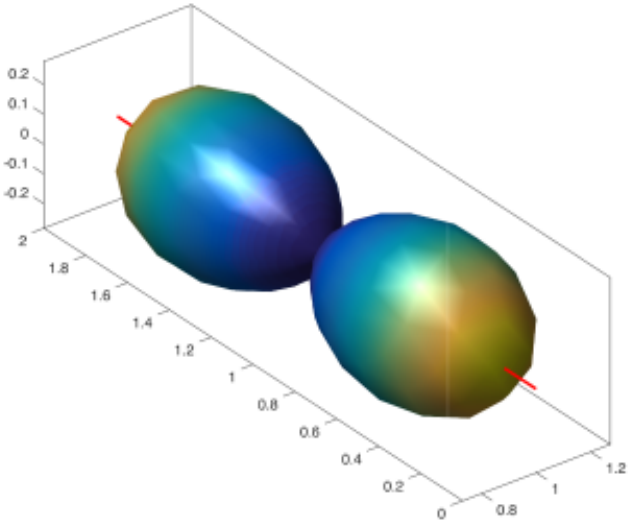}}\\
  \subfloat[HARDI
  sample]{\label{subfig:blob}\includegraphics[width=0.45\columnwidth]{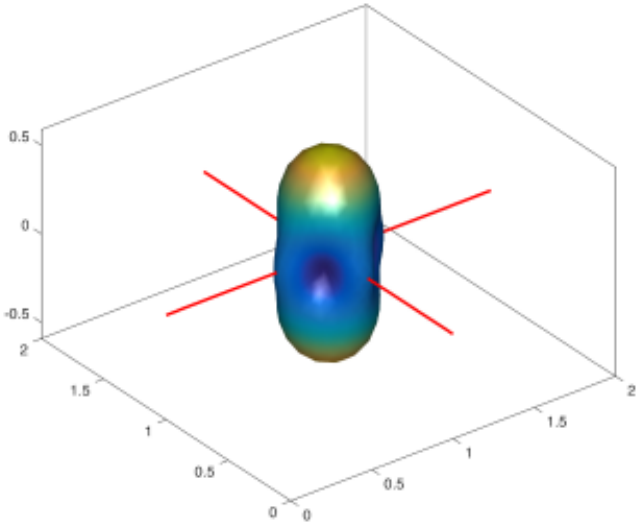}}
  \subfloat[ODF of
  (c)]{\label{subfig:cross}\includegraphics[width=0.45\columnwidth]{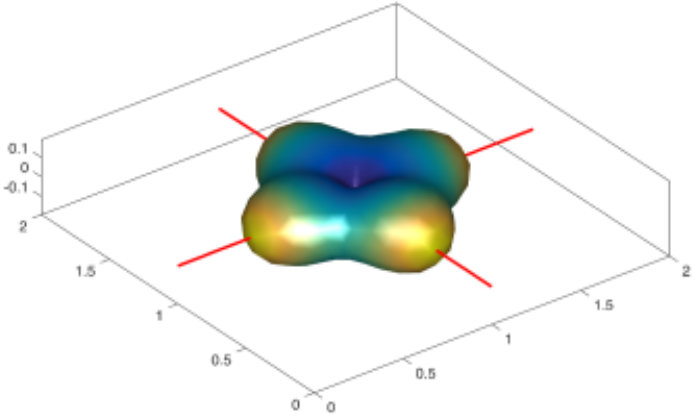}}
  \caption{Simulated DWI samples. The left column shows the raw DWI signal. The right
    column shows the reconstructed diffusion ODFs that follow anisotropic diffusion. The
    red lines indicate fiber orientations.}
  \label{fig:hardiandodf}
\end{figure}

\begin{figure}[!t]
  \centering
  \hspace*{-0cm}\subfloat[HARDI fiber tract crossing]{\label{subfig:crossnoise}\includegraphics[width=1\columnwidth]{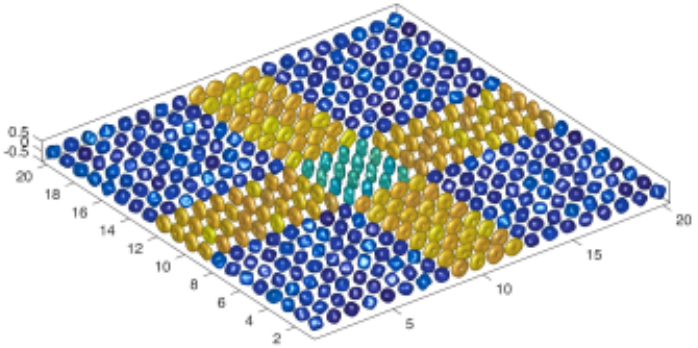}}\\
  \hspace*{-0cm}\subfloat[ODFs of (a) showing the
  fibers]{\label{subfig:crossnoiseFRT}\includegraphics[width=1\columnwidth]{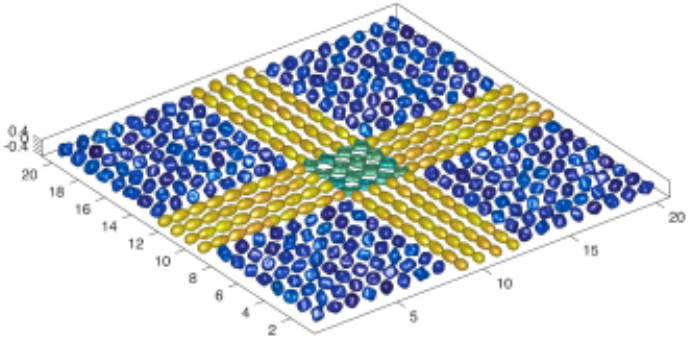}}
  \caption{Simulated DWI fiber tract crossing with uniform random noise.}
  \label{fig:crossingandnoise1}
\end{figure}

The simulated DWI (HARDI) samples were created by deforming a unit sphere of equally distributed directions~\cite{semechko2012suite} to a certain HARDI or ODF shape. \Cref{fig:hardiandodf} shows two simulated HARDI samples and their corresponding ODFs; the first is a single fiber ODF, and the second is a crossing fiber ODF. DWI samples are antipodal symmetric and every ODF from the simplest to the most complex can be constructed through a combination of single fiber ODFs. The simulated data is visualized using the regularized q-ball imaging (QBI) algorithm, which uses a linear combination of real spherical harmonics to represent either the direct QBI sample or transformed ODF \cite{descoteaux2007regularized}.
These models can be combined to form simulated DWI tracts in various DWI shapes, such as crossing fiber tracts \Cref{fig:crossingandnoise1}. A $20\times20$ grid is used throughout this section to create blueprints of fiber tract constellations to apply LORD. The images are colored according to the generalized FA (GFA) value where dark blue regions represent free isotropic diffusion. To simulate a more realistic DWI scenario, random uniform noise has been added to the samples in \Cref{fig:crossingandnoise1}.
The simulated voxels with unit density are rescaled for the free diffusion to have a low density or mean diffusivity to resemble real $b_0$ normalized data.
\begin{figure}[!t]
  \centering \subfloat[HARDI fiber tract
  crossing]{\label{subfig:crossscaled}\includegraphics[width=0.475\columnwidth]{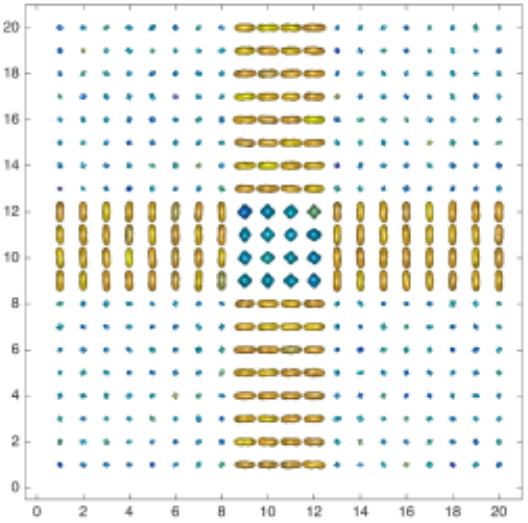}}
  \hspace*{.5cm}\subfloat[ODFs of (a) showing the
  fibers]{\label{subfig:crossscaledFRT}\includegraphics[width=0.475\columnwidth]{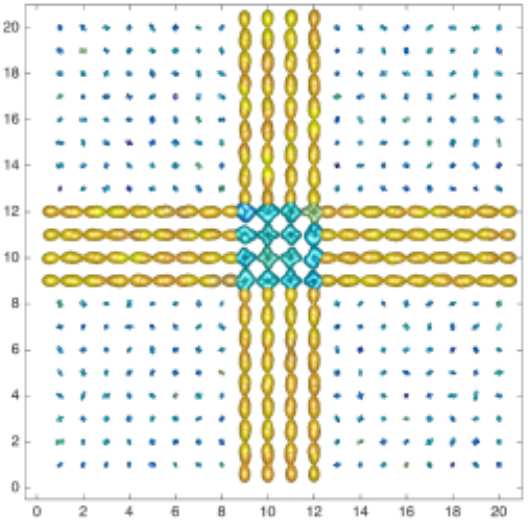}}
  \caption{Simulated DWI fiber tract crossing. Isotropic diffusion
    has now been normalized to have a lower mean diffusivity.}
  \label{fig:crossingandnoise3}
\end{figure}
\subsubsection{Parametric Setup}
For consistency, the same setup is used for all experiments on the simulated data, unless specifically stated otherwise.

\begin{description}
\item[Hierarchical mesh resolution.] In the B-spline deformation, the spacing between the control points is decreased in order to iteratively increase the degrees of freedom in the registration. We  use 4 spacings $\delta_1=4$, $\delta_2= 3.5$, $\delta_3 = 3$ and $\delta_4 = 2$. 10 iterations are used at all resolutions except the last, which terminates based on the optimal tolerance of $\epsilon = 10^{-6}$, or 90 iterations.
\item[Watson concentration, directional resolution.] The concentration parameter is set to $\kappa=15$, which is sufficiently smooth to represent the 100 uniform directions used.
\item[Spatial resolution.] A full spatial resolution is used with a B-spline smoothing at a near-Gaussian variance of $\sigma=0.6$ voxels.
\item[Histogram size.] Due to the relatively small data sample, a low number of bins is used for the histogram (i.e. intensity resolution) $20\times20$. This allows for some larger, more stable, but less refined deformations~\cite{darknersporring2012pami}.
\item[HARDI registration, ODF visualization.]\hspace{-2mm}We register the raw HARDI models, but we visualize the ODFs of the warped data based on the Funk-Radon transform (FRT). We do this to illustrate that the warped raw data is correctly reoriented and \textit{do not} visibly suffer from affine shearing.
\item[Regularization.] We set the weight $\lambda$ to $10^{-4}$. 
Each of the  examples below could be highly improved by an experiment-specific choice of parameters. However, for consistency, the same set of parameter was used for all experiments.
\end{description}

\subsection{Experiment 1: Single Fiber Tracts}
\label{subsec:singlefiber}
The first set of experiments is based on artificially generated distributions of HARDI shells and corresponding ODFs, each representing different fiber constellations. The experiment maps a straight fiber tract to a curved tract of the same width. This allows us to discuss some of the differences between a good reconstruction and a correct mapping.

\subsubsection{Straight and curved}

\begin{figure}[!ht]
  \centering
  \vspace*{-0.4cm}
  \subfloat[Moving Image]{\label{subfig:straigt_l}\includegraphics[width=0.475\columnwidth]{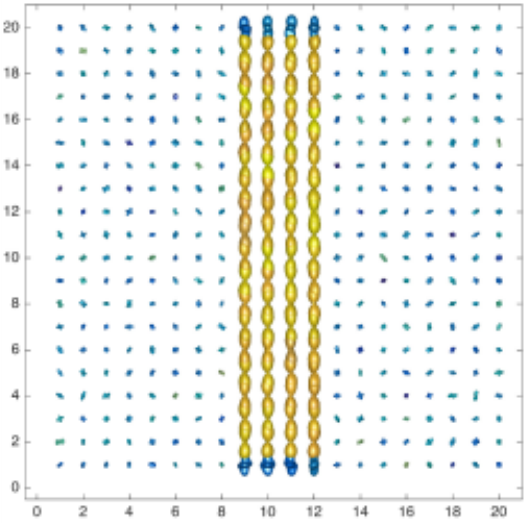}}  
  \hspace*{.5cm}\subfloat[Target Image]{\label{subfig:wavy_l}\includegraphics[width=0.475\columnwidth]{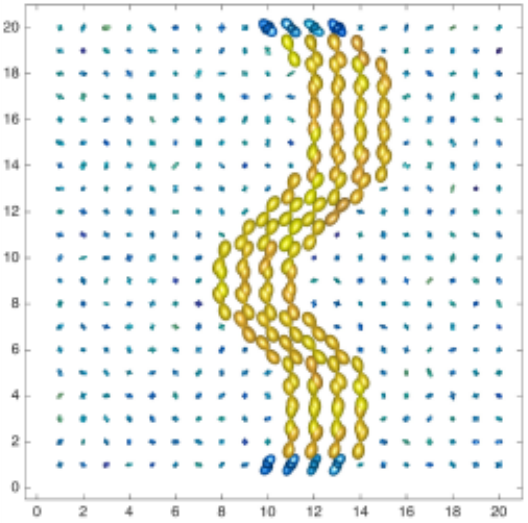}} 
  \caption{Experiment 1: Simulated fiber tract images with intersecting fibers at the boundaries.
  }
  \label{fig:straight2wavy_l}
\end{figure}

\begin{figure}[!ht]
  \centering \vspace*{-0.4cm} \subfloat[Reconstructed warped
  image]{\label{subfig:straigt2wavy_deform}\includegraphics[width=0.475\columnwidth]{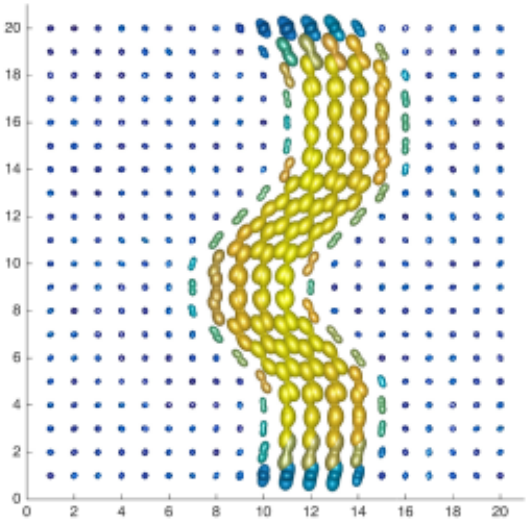}}
  \hspace*{.5cm}\subfloat[Spatial
  mapping]{\label{subfig:straigt2wavy_arrows}\includegraphics[width=0.475\columnwidth]{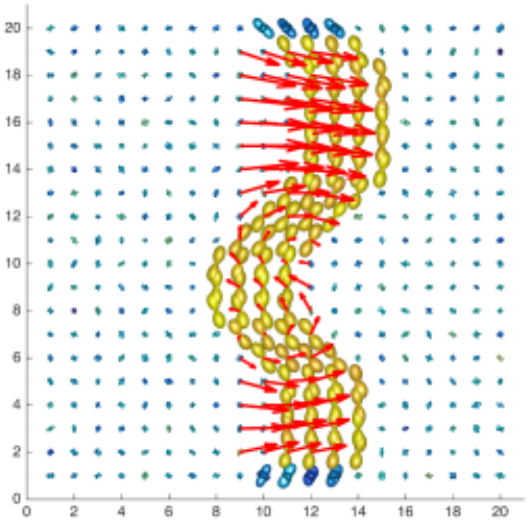}}
  \caption{Experiment 1: Registration of single tract images of varying shape and (given
    the boundary) length.}
  \label{fig:straight2wavy_l_results}
\end{figure}

\begin{figure}[!ht]
  \centering \vspace*{-0.4cm} \subfloat[Reconstructed warped
  image]{\label{subfig:straigt2wavy_deform_boundary_k0}\includegraphics[width=0.475\columnwidth]{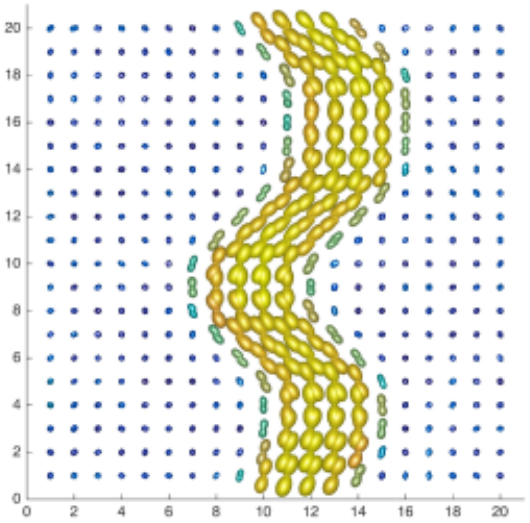}}
  \hspace*{.5cm}\subfloat[Spatial
  mapping]{\label{subfig:straigt2wavy_arrows_boundary_k0}\includegraphics[width=0.475\columnwidth]{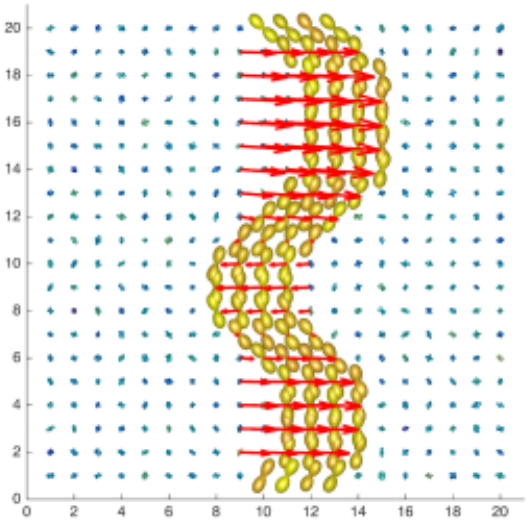}}
  \caption{Experiment 1: Scalar-based registration without boundary fibers and no directional information in the cost function during optimization.}
  \label{fig:straight2wavy_l_boundary_k0}
\end{figure}

Three experiments illustrate different scenarios where straight fibers are registered to wavy fibers. These experiments illustrate the regularizing effects of using the full diffusion profile for registration.  In the first experiment, the position of the end of the fibers is constrained by adding intersecting fibers at the end and beginning. \Cref{fig:straight2wavy_l} shows the simulated moving (straight) and target (wavy) images.
The results of the first experiment are shown in \Cref{fig:straight2wavy_l_results}, where \Cref{subfig:straigt2wavy_deform} is the reconstructed warp, and \Cref{subfig:straigt2wavy_arrows} shows the final spatial mapping from the moving image, overlaid on the original target image. As the figures show, the straight fibers are stretched in correspondence with the curvature, and the reconstructed ODFs from the HARDI-based registration are rotated correctly, although smoothed by the interpolation.
\begin{figure}[!ht]
  \centering \vspace*{-0.4cm} \subfloat[Reconstructed warped
  image]{\label{subfig:straigt2wavy_deform_boundary}\includegraphics[width=0.475\columnwidth]{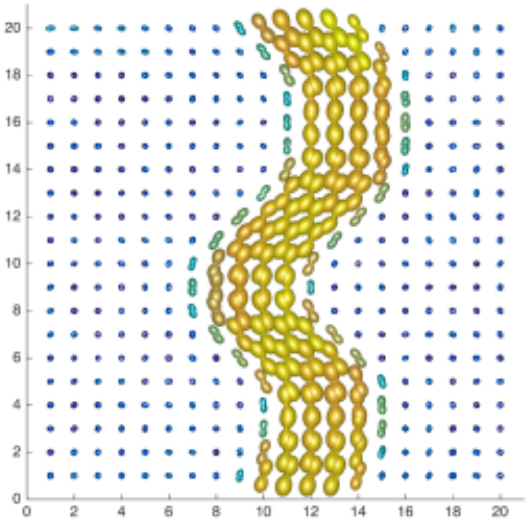}}
  \hspace*{.5cm}\subfloat[Spatial
  mapping]{\label{subfig:straigt2wavy_arrows_boundary}\includegraphics[width=0.475\columnwidth]{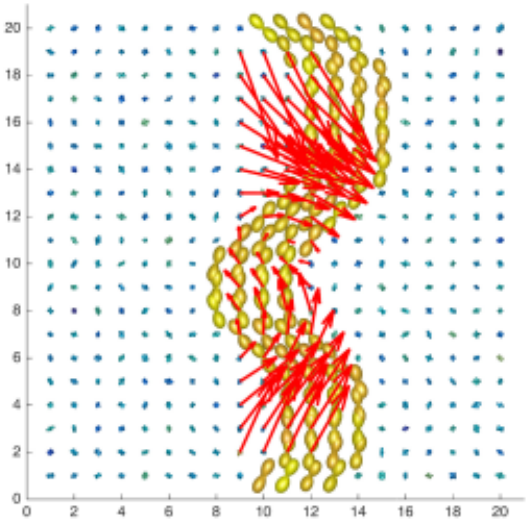}}
  \caption{Experiment 1: Registration of without boundary fibers The spatial mapping
    indicates a nice preservation of length of the straight tract mapped to a subsection
    of the curved tract.}
  \label{fig:straight2wavy_l_boundary}
\end{figure}
The second experiment is performed without features on the boundaries. The results are shown in \Cref{fig:straight2wavy_l_boundary}, where we observe that the length is preserved as the straight fiber tract is mapped to a sub-part of the curved tract. Intuitively, a correct registration in such a case should preserve the length of the straight fiber after deformation.
The fact that this happens with no strong outside regularization, \revchange{seems to indicate an inherent regularization effect} of the cost function.  In the third experiment, the proposed similarity is compared with the equivalent scalar-based registration by performing a mean diffusivity registration ($\kappa=0$) of the tracts with no signal on the boundaries. The mean diffusivity carries no directional information. The result are shown in \Cref{fig:straight2wavy_l_boundary_k0}.
This pure scalar-based registration is driven by the edges of the simulated tracts only. The reconstruction appears to be correct, but the final spatial mapping indicates a lack of regularization as the fibers are stretched unevenly.

\subsection{Experiment 2: Crossing Fiber Tracts}
\label{subsec:crossingfiber}

The second set of experiments are designed to investigate the registration of crossing fiber tracts.

\subsubsection{Straight and shifted}

\begin{figure*}[!t]
  \centering \subfloat[Moving
  Image]{\label{subfig:straight_x_2shift}\includegraphics[width=0.5\columnwidth]{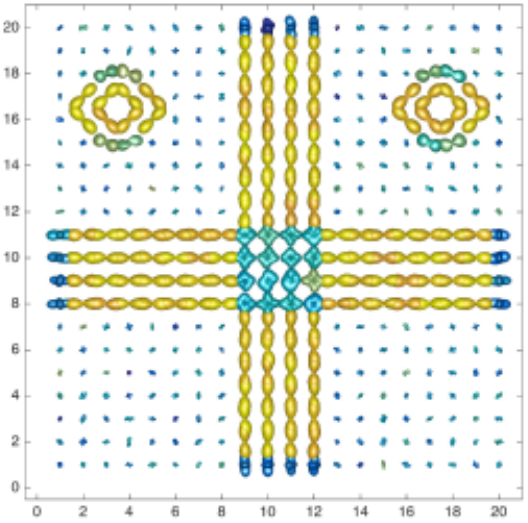}}
  \hspace*{.1cm}\subfloat[Target
  Image]{\label{subfig:straight_x_shifted}\includegraphics[width=0.5\columnwidth]{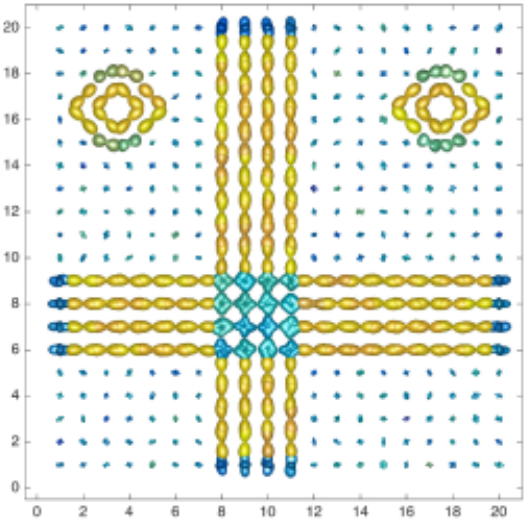}}
  \hspace*{.1cm}\subfloat[Reconstructed warped
  image]{\label{subfig:straigt2shift_deform}\includegraphics[width=0.5\columnwidth]{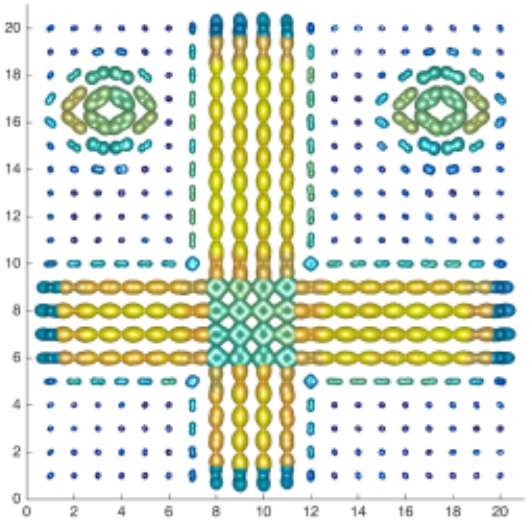}}
  \hspace*{.1cm}\subfloat[Spatial
  mapping]{\label{subfig:straigt2shift_arrows}\includegraphics[width=0.5\columnwidth]{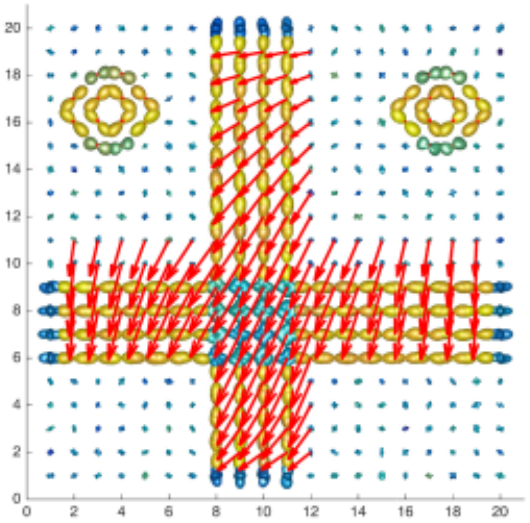}}
  \caption{Experiment 2: Simulated crossing tracts (a) with a vertical shift (b). The
    registered result of the crossing under both vertical and horizontal shift is
    reconstructed in (c), and shown with the final spatial mapping from the original
    position of the moving image in (d).}
  \label{fig:straight2shift_x_results}
  \vspace*{-0.5cm}
\end{figure*}

The first experiment examines the framework's ability to register two crossing tracts with a horizontal and vertical shift \Cref{subfig:straight_x_2shift,subfig:straight_x_shifted}. Circular fibers have been added as fixed points in the image to illustrate the local shift of the crossing tracts. The result of the registration is shown in \Cref{subfig:straigt2shift_deform,subfig:straigt2shift_arrows}. The final spatial mapping, shown with arrows, is accurately including the reconstruction. Note that the reconstruction is subject to smoothing effect in the interpolation.

\subsubsection{Two degrees of shearing}

\begin{figure*}[!t]
  \centering \subfloat[Moving
  Image]{\label{subfig:sheared30}\includegraphics[width=0.5\columnwidth]{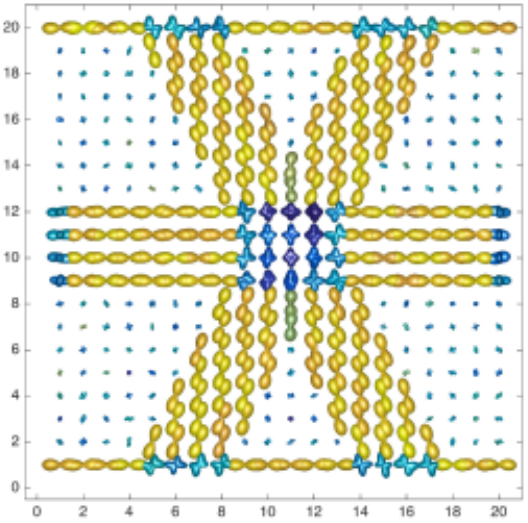}}
  \hspace*{.1cm}\subfloat[Target
  Image]{\label{subfig:sheared40}\includegraphics[width=0.5\columnwidth]{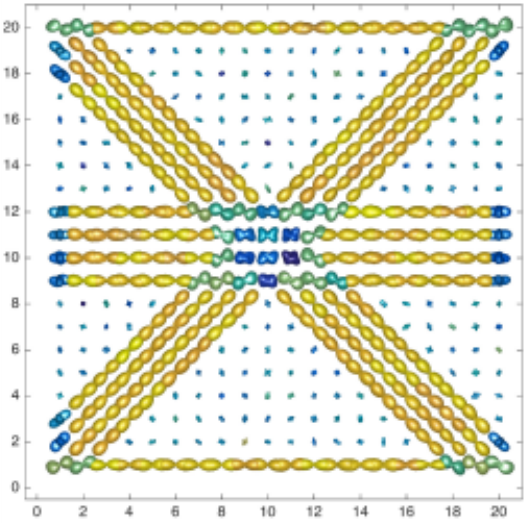}}
  \hspace*{.1cm}\subfloat[Reconstructed warped
  image]{\label{subfig:sheared_deform}\includegraphics[width=0.5\columnwidth]{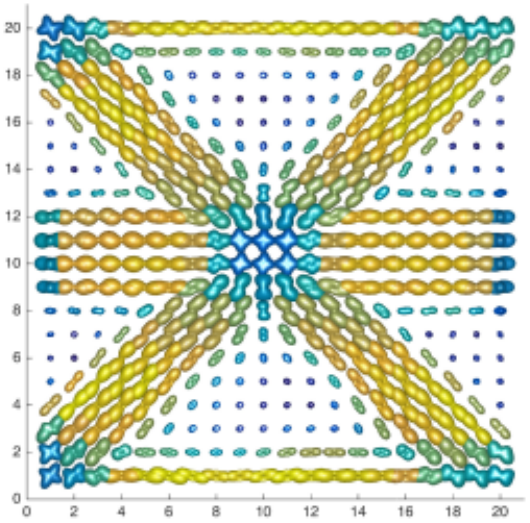}}
  \hspace*{.1cm}\subfloat[Spatial
  mapping]{\label{subfig:sheared_arrows}\includegraphics[width=0.5\columnwidth]{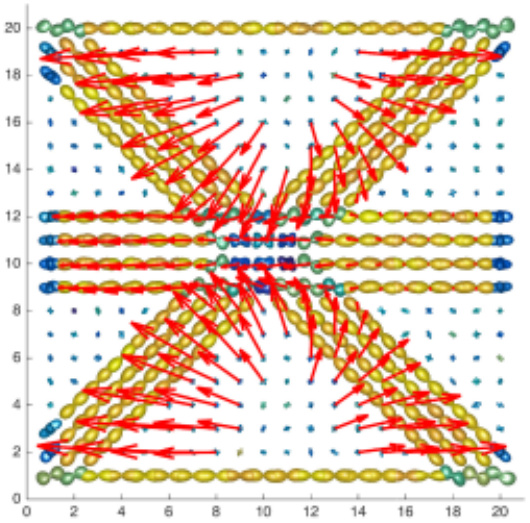}}
  \caption{Experiment 2: Simulated crossing tracts with $\sim$30 degree (a) to 45 degree
    shearing (b). The reconstruction of the registered result is shown in (c), and the
    corresponding spatial mapping in (d).}
  \label{fig:sheared2sheared_x_results}
  \vspace*{-0.5cm}
\end{figure*}

The second experiment involving crossing fibers shows three fiber tracts crossing under a varying amount of shear with a fixed horizontal tract (\Cref{subfig:sheared30,subfig:sheared40}). The purpose is to investigate a relatively large deformation combined with a change to the complex crossing at the center The results are shown in \Cref{subfig:sheared_deform,subfig:sheared_arrows}. As the figure shows, the structure of the complex center-crossing closely matches the orientations of 45 degree crossing fibers. We remind the reader that the registration was not performed on the ODFs but directly on the simulated HARDI models and subsequently reconstructed.

\subsection{Experiment 3: Fanning Fiber Tracts}
\label{subsec:kissingfiber}
The third set of experiments examines the most complex configurations namely the registration of kissing and fanning fibers. Note that the artificial cases constructed in these experiments are such that moving and target images are in a 1-1 correspondence.
\subsubsection{Fanning and kissing fiber tracts in a crossing}
\begin{figure*}[!t]
  \centering \subfloat[Moving
  Image]{\label{subfig:fanning}\includegraphics[width=0.5\columnwidth]{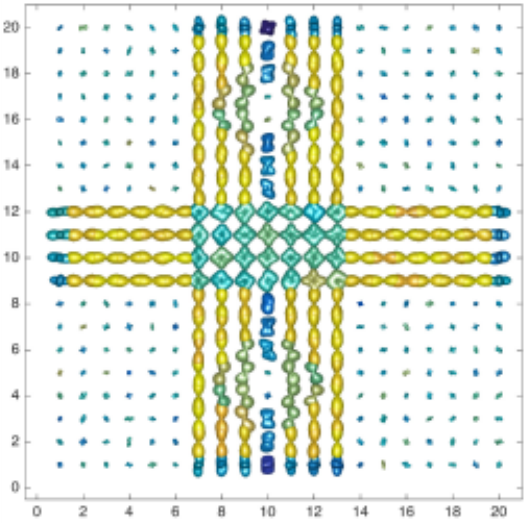}}
  \hspace*{.1cm}\subfloat[Target
  Image]{\label{subfig:kissing}\includegraphics[width=0.5\columnwidth]{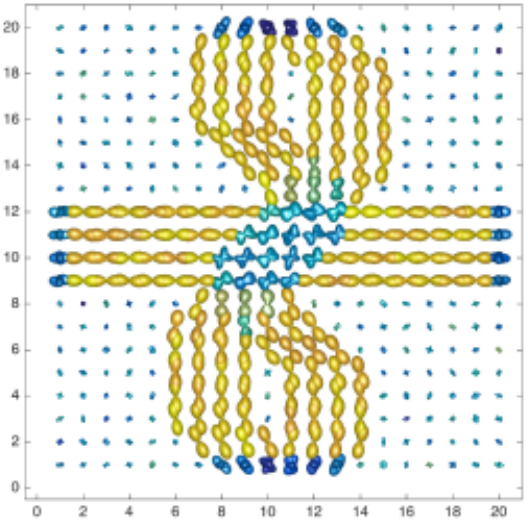}}
  \hspace*{.1cm}\subfloat[Reconstructed warped
  image]{\label{subfig:fanning_deform}\includegraphics[width=0.5\columnwidth]{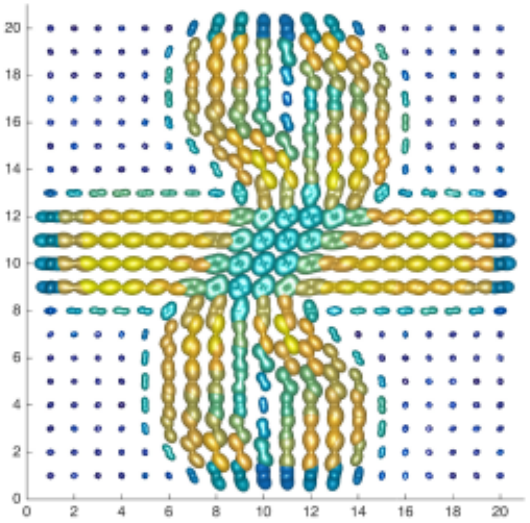}}
  \hspace*{.1cm}\subfloat[Spatial
  mapping]{\label{subfig:kissing_arrows}\includegraphics[width=0.5\columnwidth]{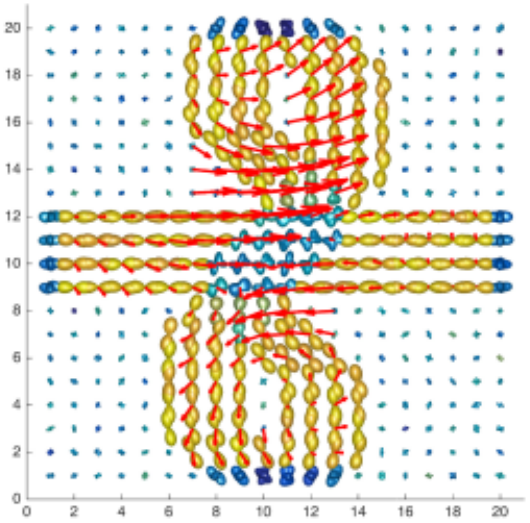}}
  \caption{Experiment 3: (a) Simulated fanned opening in the moving image, and (b) an
    interleaving curved target image. The reconstruction is shown in (c), and the spatial
    mapping in (d).}
  \label{fig:fanning2kissing_x_results}
  \vspace*{-0.5cm} 
\end{figure*}

The first experiment consist of two DWI images, that simulate both fanning (dispersing) and kissing (interleaving) fiber tracts. The moving image in \Cref{subfig:fanning,subfig:kissing} contains a crossing with a few fibers fanning in and out along the vertical centerline. The target image contains two curved tracts fanning in and out, and merging at the central crossing. The results of the registration experiment are shown in \Cref{subfig:fanning_deform,subfig:kissing_arrows}, where both the reconstructed warp and spatial mapping show a registration that follows the lines of the original target image. The resulting registration can move, shrink, and turn the center-crossing to fit the target image.
\subsubsection{Kissing fiber tracts}

\begin{figure*}[!t]
  \centering \subfloat[Moving
  Image]{\label{subfig:straightlines}\includegraphics[width=0.5\columnwidth]{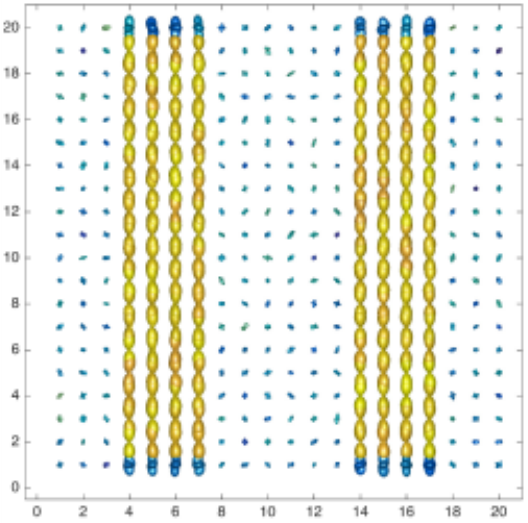}}
  \hspace*{.1cm}\subfloat[Target
  Image]{\label{subfig:kissinglines}\includegraphics[width=0.5\columnwidth]{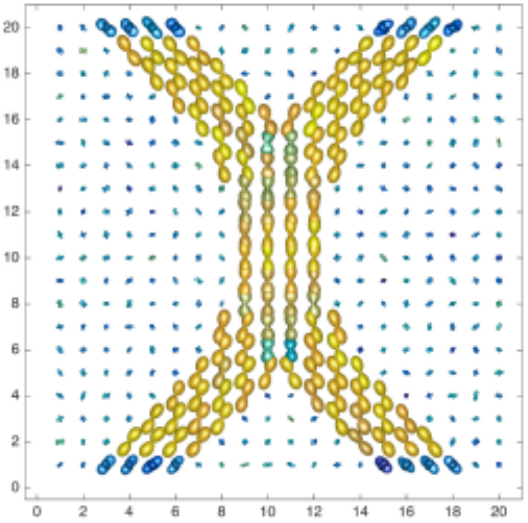}}
  \hspace*{.1cm}\subfloat[Reconstructed warped
  image]{\label{subfig:straight2kiss_deform}\includegraphics[width=0.5\columnwidth]{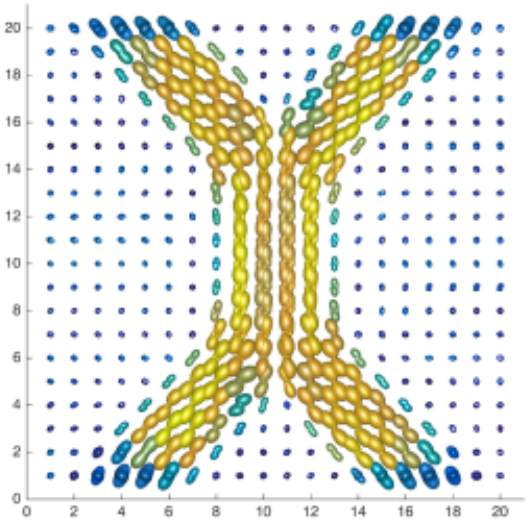}}
  \hspace*{.1cm}\subfloat[Spatial
  mapping]{\label{subfig:straight2kiss_arrows}\includegraphics[width=0.5
    \columnwidth]{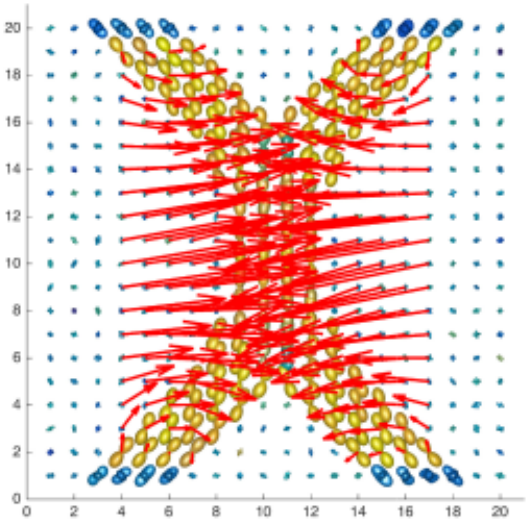}}
  \caption{Experiment 3: Simulated straight (a) and kissing (b) fiber tracts. The
    reconstruction is shown in (c), and the spatial mapping in (d).}
  \label{fig:straight2kiss_results}
\end{figure*}

The second experiment involves two straight fiber tracts that are registered to two curving and interleaving tracts (\Cref{subfig:straightlines,subfig:kissinglines}).  The result, shown in \Cref{subfig:straight2kiss_deform,subfig:straight2kiss_arrows}, displays a large and difficult deformation that by all accounts appears to be a successful registration.


\subsection{Synthetic Deformations of Real Data}
\label{sec:syndeform}

In this set of experiments, the registration framework is evaluated on real data obtained from the HCP \cite{van2013wu}. By introducing a random synthetic warp on a subset of the brain data, ground truth is obtained where the goal is to register the warped image back to the original image. These experiments illustrate the effect of the parameters of the model in a realistic, but guaranteed  diffeomorphic scenario.
\subsubsection{DWI example data}

\begin{figure}[H]
  \centering
  \includegraphics[width=1\columnwidth]{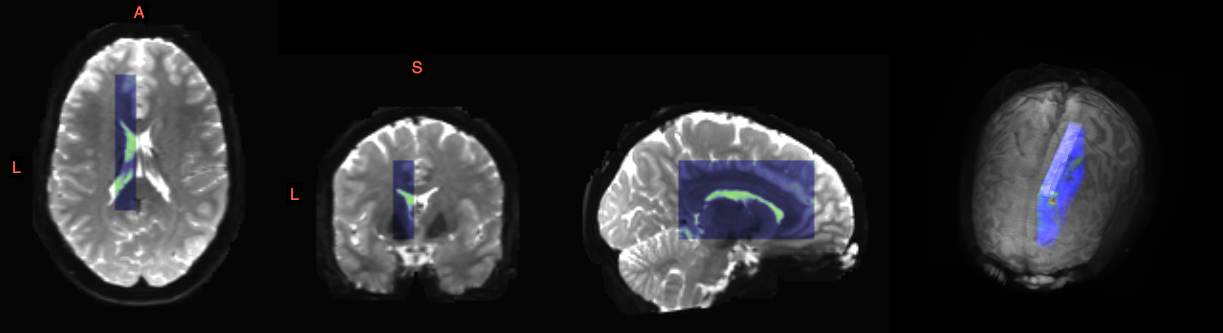}
  \caption{Selected region of interest for synthetic warp experiments (blue) overlaid on
    the $b_0$ image of HCP subject \texttt{103818}.}
  \label{fig:cc_roi_b0}
\end{figure}
\begin{figure}[ht]
  \centering
  \subfloat[Sagittal slice at ROI center.]{\label{subfig:cc_slice}\includegraphics[width=1\columnwidth]{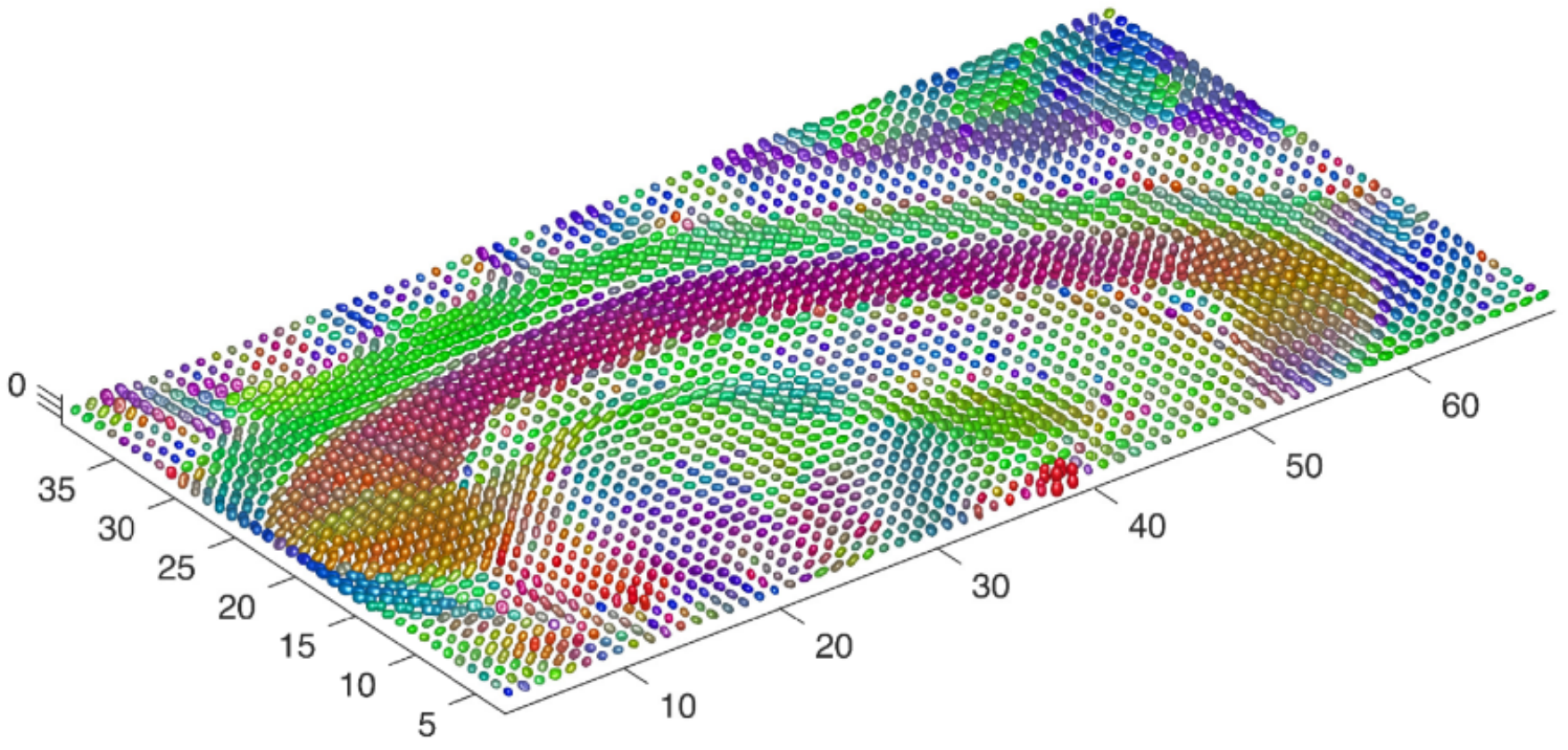}}\\
  \subfloat[Deformation
  field.]{\label{subfig:cc_deform}\includegraphics[width=1\columnwidth]{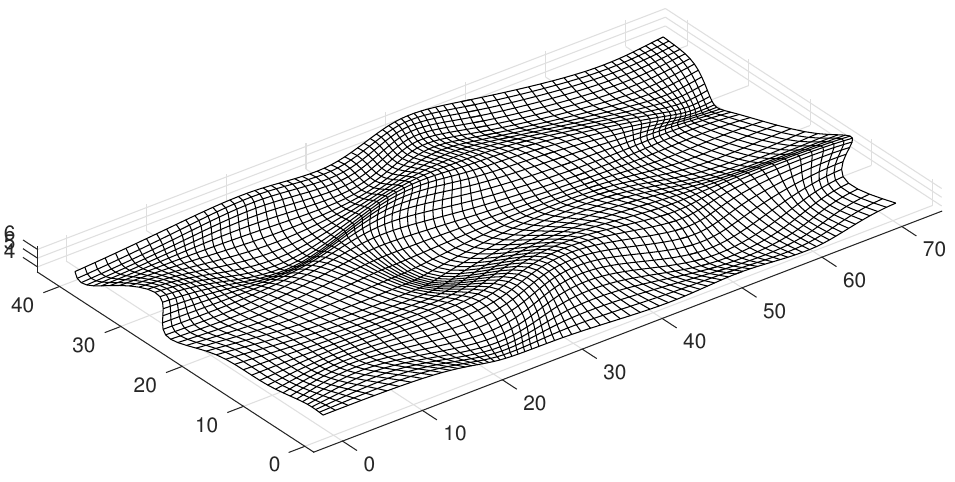}}
  \caption{Original central ROI slice with ODFs (a), and the deformation to be applied
    (b).}
  \label{fig:cc_slice2}
\end{figure}

The DWI data used in this experiment is shown in \Cref{fig:cc_roi_b0}, where the region of interest (ROI) is the blue overlay on the subject's $b_0$ image. An ROI of the brain is used to improve the visualization of the results. The ROI was chosen at the edge of the corpus callosum (CC) in the left hemisphere due to the characteristic C-shape of the CC and the intersection with other well-known structures e.g. the cingulum. Furthermore, the ROI is in an area with crossing fiber tracts and is near the cortex. A $b=1000$ DWI volume is used with  a ROI size $11\times71\times41$ at 1.25mm isotropic voxels with 90 directions. Only the central sagittal slice along with corresponding ODFs is visualized (\Cref{subfig:cc_slice}), while the deformation is applied to the whole ROI. The deformation field for the central slice is shown in \Cref{subfig:cc_deform}.

\subsubsection{Parameters effect}
The experimental setup provides a ground truth: the identity transformation. As the similarity measure, NMI, does not reflect a unique point-wise correspondence, we introduce three measures for evaluating the resulting registration: (1) mean squared error (MSE) between coordinates, (2) curl of the final deformation field, and (3) divergence of the final deformation field \cite{helmholz1858}.

\textit{Curl}
quantifies the amount of orientational change in the deformation field, also referred to as rotation or vorticity. We use the $L_2$-norm of the curl vector, i.e. the rotation velocity. A higher magnitude of the curl is expected for the results of a scalar-based registration compared to a registration over the full diffusion profile. 

\textit{Divergence} quantifies the density of the outward flux of a vector field which can be either positive or negative, indicating an expansion or a contraction at a given point. In combination, these three measures provide information about the magnitude of the voxel-wise distance and the state of the final deformation field. Curl and divergence have previously been used as regularization in a nonrigid registration framework \cite{riyahi2014regularization}.

We  report now results of experiments where we varied the intensity scale, the orientation scale, and the spatial scale: the intensity and orientational scales are the elements which distinguish LORD from the other registration frameworks. We investigated the following:
\begin{description}
\item[1. Isointensity curves.] The density-based formulation\linebreak allows us to smooth the image inversely proportionally to image gradients, such as borders near the cerebrospinal fluid region, which can be strongly affected by partial volume effects. We iteratively decrease the control point-spacing in the FFD model to get an increasingly localized registration and increase the intensity range accordingly. This is realized with an initial histogram with relatively few bins and a fixed degree of smoothing and then successively increase the number of bins as the deformation resolution is increased.
\item[2. Explicit reorientation.] The directional information is expected to provide a more stable and regularized transformation. To investigate this hypothesis, different levels of directional smoothing are examined from $\kappa=30$,  a highly peaked ODF, to a scalar-based mean diffusivity registration, i.e. setting the concentration parameter $\kappa$ to $0$. The experiments on simulated data indicated that the directional information results in direct regularized solutions through the similarity, and these experiments seek to uncover if these observations hold for real data.
\end{description}
Performing multi-resolution registration (a continuation method) is a key element in most high resolution registration frameworks, e.g. in the FFD model \cite{rueckert1999nonrigid}, \texttt{ANTs} \cite{avants2009advanced}, \texttt{Elastix} \cite{klein2010elastix}, \texttt{FSL} \cite{jenkinson2012fsl}, etc. It provides stability while allowing for large and small deformations and reducing the computational complexity. While experiments regarding the spatial part will be performed, its impact on registration quality is fairly well-described in contrast to iso-intensity curves and explicit reorientation.

\subsubsection{Parametric Setup}

\begin{description}
\item[Hierarchical mesh resolution.] Similar to the simulated experiments, the spacing between the control points is successively decreased. To this end we use the composition $\delta_1=10$, $\delta_2=5$, $\delta_3=3.5$, $\delta_4=3$, (the control point spacing $\delta$ is scaled down according to the dimensions of the image). The low initial resolution of the deformation permits the generation of a large random and valid synthetic warp at around half of the control point spacing ($\sim$ $0.4\delta$) \cite{rueckert2006diffeomorphic}. For the optimization, we use a quasi-Newton L-BFGS optimizer which runs for 50 iterations for all resolutions, unless the optimality condition of $\epsilon = 10^{-6}$ is fulfilled. Note that we refer to the result after each optimizer termination as \textit{a step}, thus 4 steps in total.
\item[Accumulated curl and divergence.] While the MSE is\linebreak  easy to calculate at each step, the curl and divergence depend on the first-order derivatives of the spatial deformation, which are accumulated over\linebreak each step. Thus the final curl and divergence is defined from the product of Jacobians for spatial resolution $Curl(\Phi_{\delta=3})\!=\! Curl(\Phi_{\delta=10}) + Curl(\Phi_{\delta=5}) +$\linebreak $Curl(\Phi_{\delta=3.5})+Curl(\Phi_{\delta=3})$ for all voxels. 
\item[Change in resolution.] Changing the spatial resolution is obtained by convolution.
\item[Other fixed parameters.] The regularization is fixed to $\lambda=10^{-4}$, and we interpolate and optimize over 30 interpolated orientations out of the 90 ones present in the HCP data.
\end{description}
All the following experiments will show the results of the four accumulating steps. All error measures are reported for the entire ROI and not just the slice visualized.

\subsubsection{The Quantitative Effect of Isoparametric Curves}
\label{subsec:isocurvesparam}

\begin{figure*}[!htb]
  \centering
  \hspace*{0cm}\subfloat[MSE]{\label{subfig:binsmse}\includegraphics[width=0.60\columnwidth]{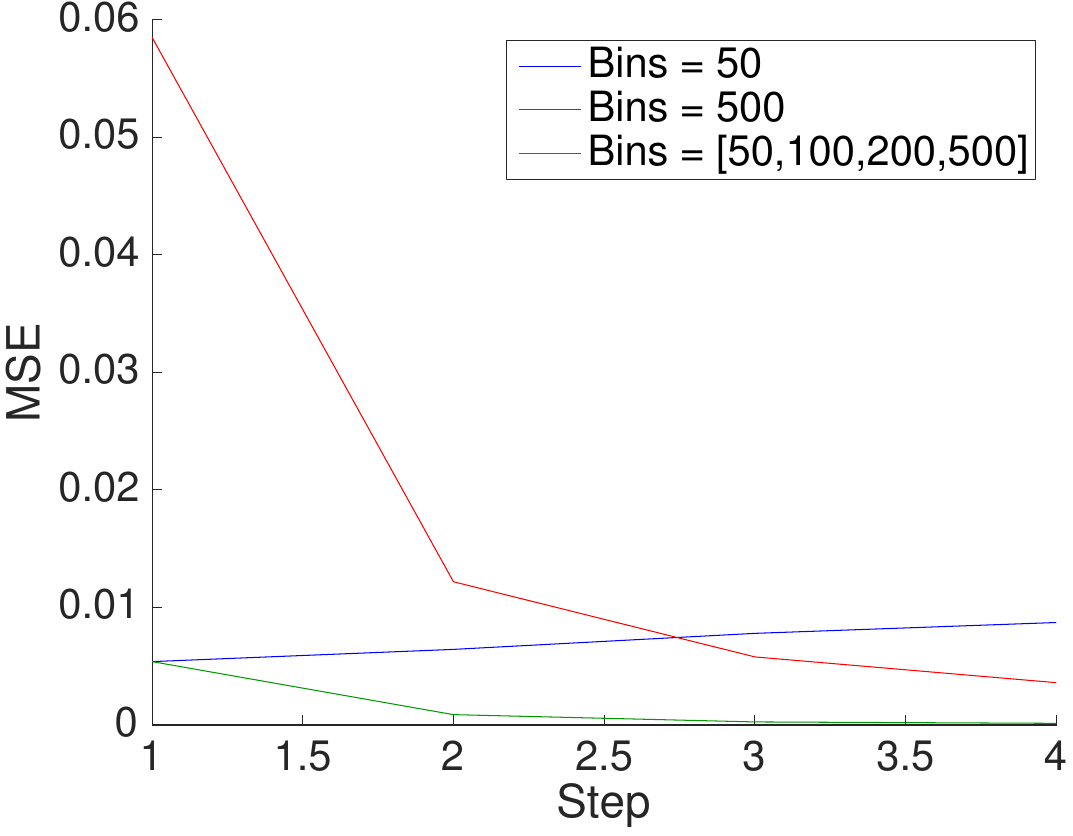}}
  \hspace*{0cm}\subfloat[Curl]{\label{subfig:binscurl}\includegraphics[width=0.60\columnwidth]{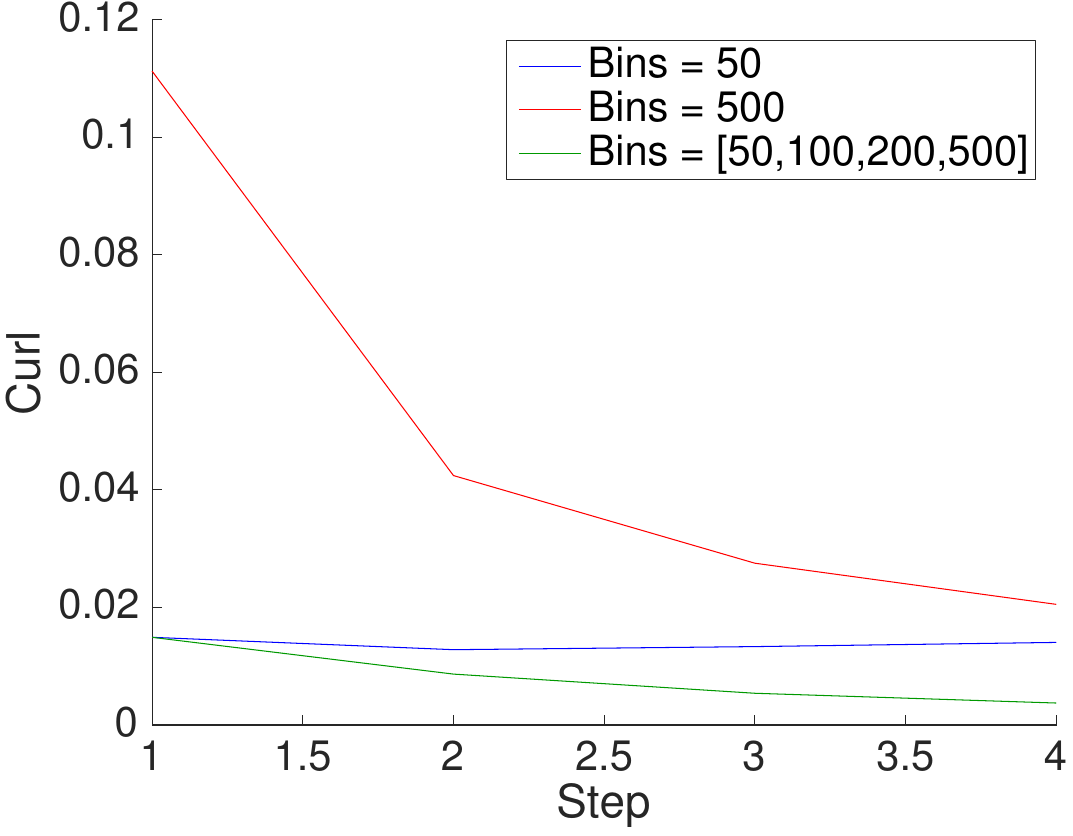}}
  \subfloat[(Absolute)
  Divergence]{\label{subfig:binsdiv}\includegraphics[width=0.60\columnwidth]{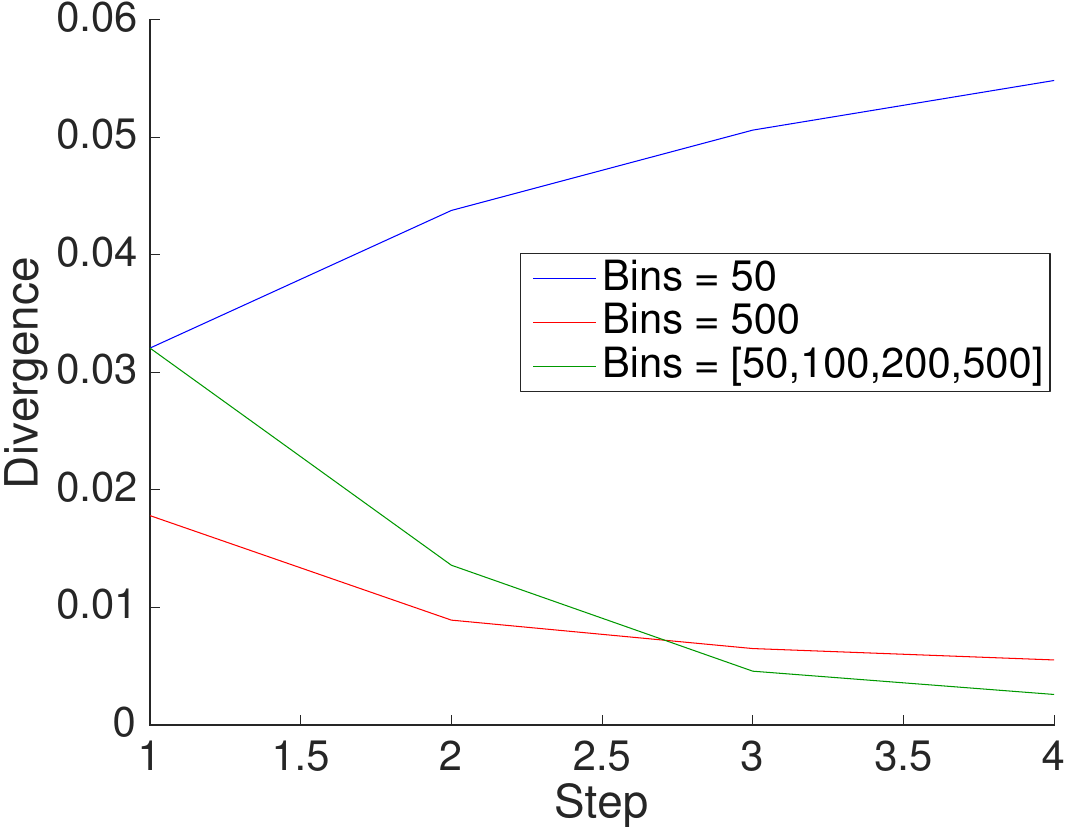}}
  \caption{Results from testing the effect of the number of bins through the four step
    registration. The lines are the mean over all points. The best results are (a): green
    at $1.42\times10^{-4}$, (b): green at $3.74\times10^{-3}$, and (c): green at
    $2.60\times10^{-3}$.}
  \label{fig:binstest}
\end{figure*}

This experiment illustrates the effect of changing the size of the smoothed joint histogram. Three experiments are performed with (i) a fixed histogram of $50\times50$ bins, (ii) a histogram with $500\times500$ bins, and (iii) gradually increasing the histogram size in 50, 100, 200, and 500 bins. \Cref{fig:binstest} shows the results in terms of MSE, curl, and divergence as mean value over all points along.
It is clear that a small histogram with wide isocurves results in an initially faster convergence \Cref{fig:binstest} (blue line). However, wide iso-curves result in flat regions with small gradients and little structure, which may cause the result to become sub-optimal or degenerate with increasing degrees of freedom in the transformation. In contrast, starting with a high bin resolution  histogram with thin isocurves has the opposite effect in the initial steps (red line). However, by iteratively refining the joint histogram, we can refine image motion, from large to small, and this provides a better result (green line).

\subsubsection{The Quantitative Effect of the Orientation Scale}
\label{subsec:orientationparam}

\begin{figure*}[!htb]
  \centering
  \hspace*{0cm}\subfloat[MSE]{\label{subfig:kappamse}\includegraphics[width=0.60\columnwidth]{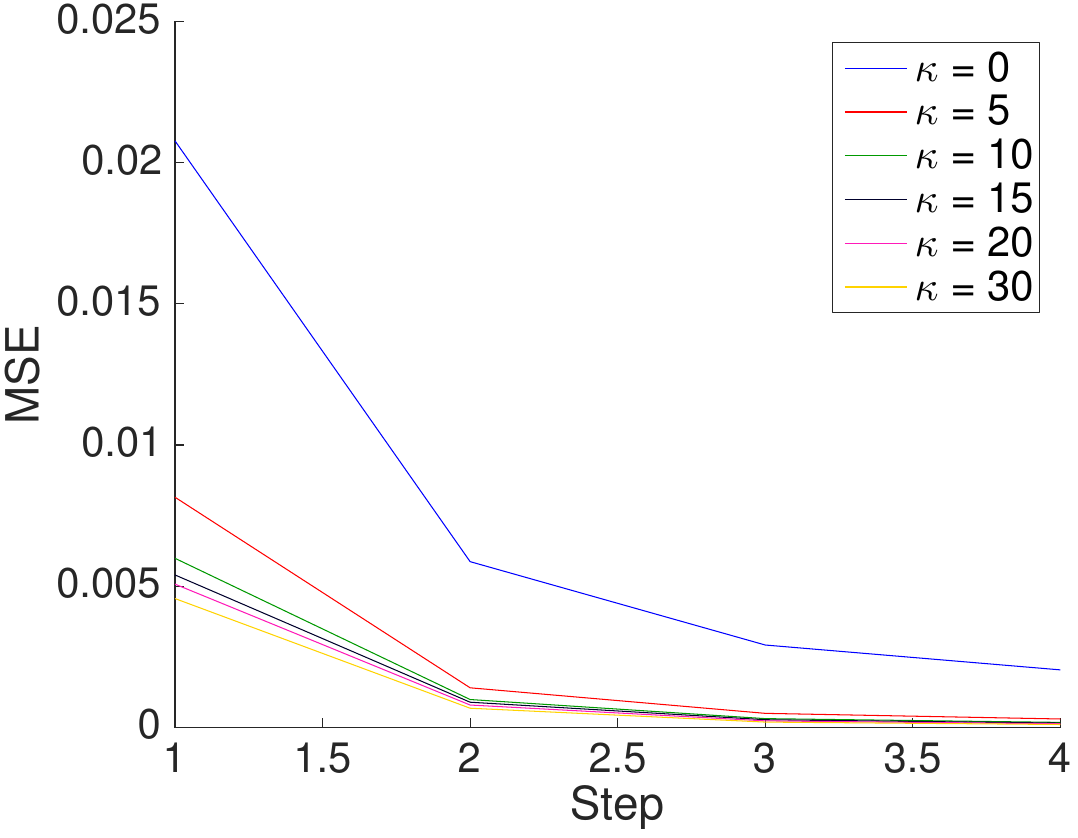}}
  \hspace*{0cm}\subfloat[Curl]{\label{subfig:kappacurl}\includegraphics[width=0.60\columnwidth]{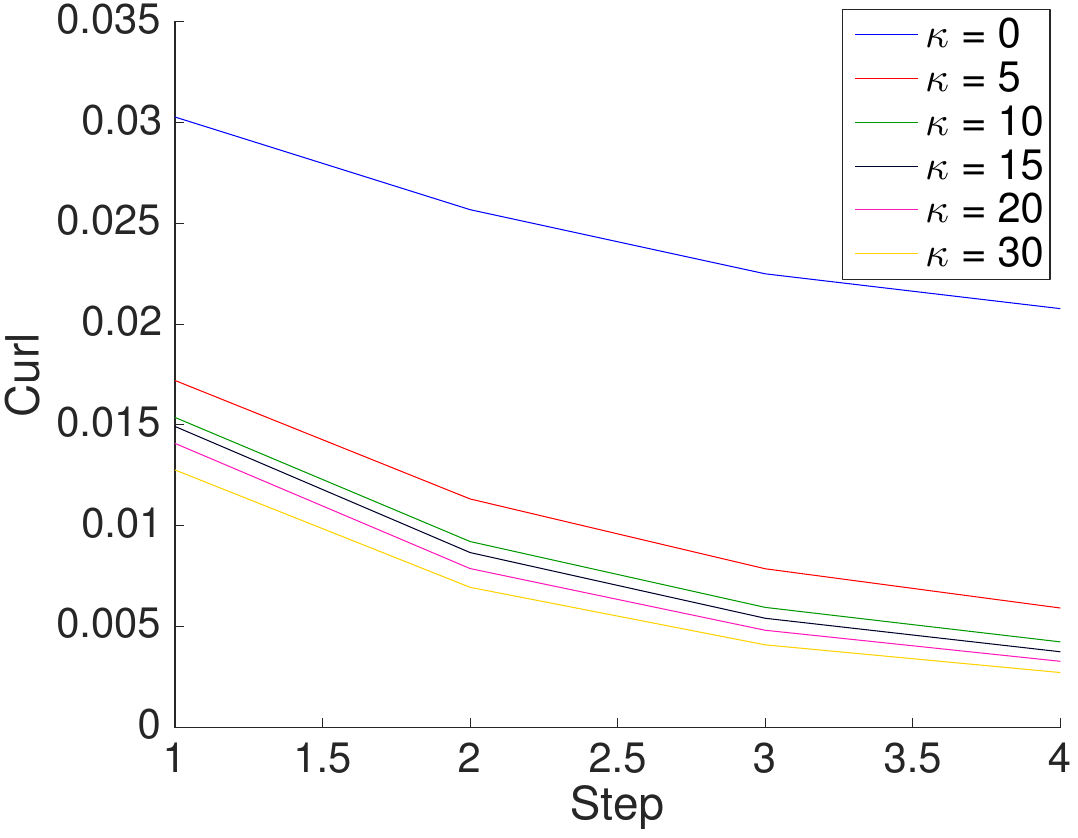}}  
  \subfloat[(Absolute) Divergence]{\label{subfig:kappadiv}\includegraphics[width=0.60\columnwidth]{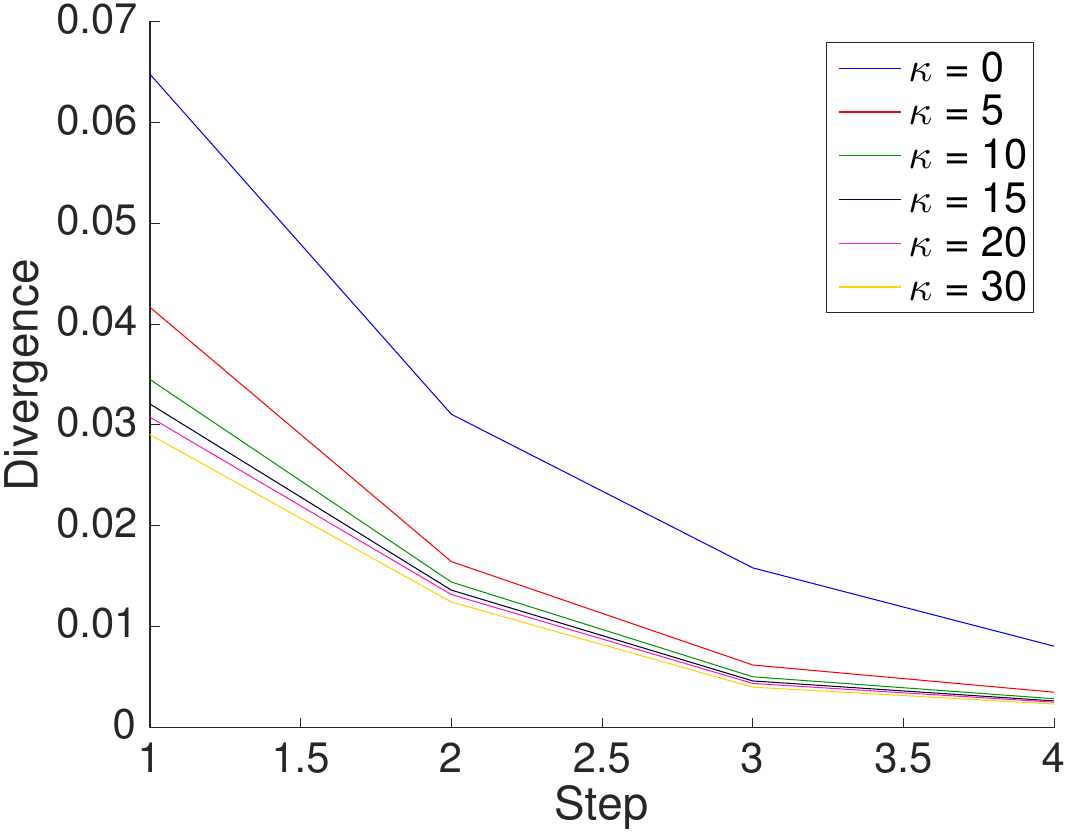}}
  \caption{Results from testing the effect of smoothness of the directional interpolation through the four-step registration. The lines are the mean overall points. The best
    results are the yellow line at $\kappa=30$ with (a): $9.17\times10^{-5}$, (b):
    $2.71\times10^{-3}$, and (c): $2.29\times10^{-3}$. However, the overall difference
    between $\kappa=10$ and $\kappa=30$ approx, $10^{-5}$.}
  \label{fig:kappatest}
\end{figure*}

In this experiment, we investigate whether directional information increases the stability and improves the registration. We employ the size progression of the histogram described in the previous experiment (i.e. $50\times 50$, $100\times100$, $200\times 200$ and $500\times 500$ bins), and we vary the concentration parameter $\kappa$ from mean diffusivity at $\kappa=0$ to sharp angular features at $\kappa=30$. The results are shown in \Cref{fig:kappatest}.
The figure shows how the directional information results in better registration, with significantly less curl than the scalar registration at $\kappa=0$ and we observe that the best value $\kappa=30$ is stable in high directional resolution data such as the HCP. We suggest setting $\kappa=15$ as this should be enough for both high and low-resolution data, such as DTI.

\subsubsection{The Quantitative Effect of the Spatial Resolution}
\label{subsec:spatialparam}

In the last experiment, we use $\kappa=15$, set again the bins to $[50\times50, 100\times100,200\times200, 500\times500]$, and investigate the effects of the spatial scale.
\begin{figure*}[!htb]
  \centering
  \hspace*{0cm}\subfloat[MSE]{\label{subfig:spatialmse}\includegraphics[width=0.55\columnwidth]{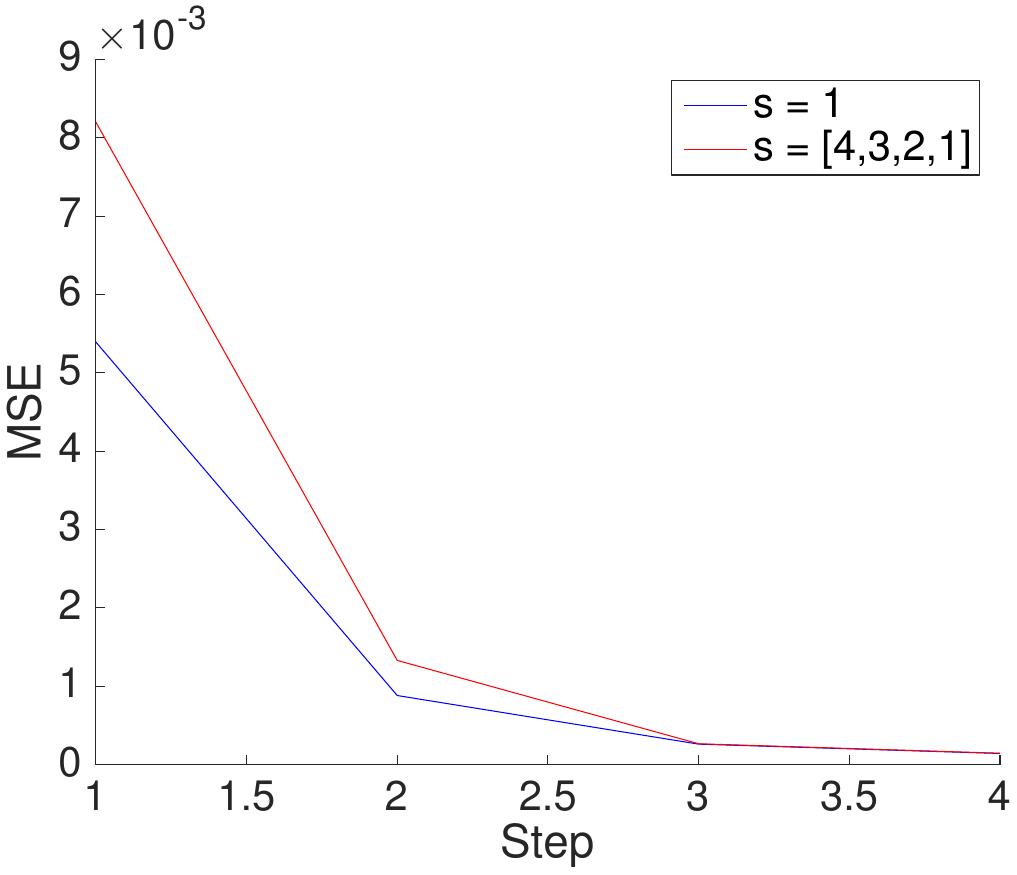}}
  \hspace*{0cm}\subfloat[Curl]{\label{subfig:spatialcurl}\includegraphics[width=0.55\columnwidth]{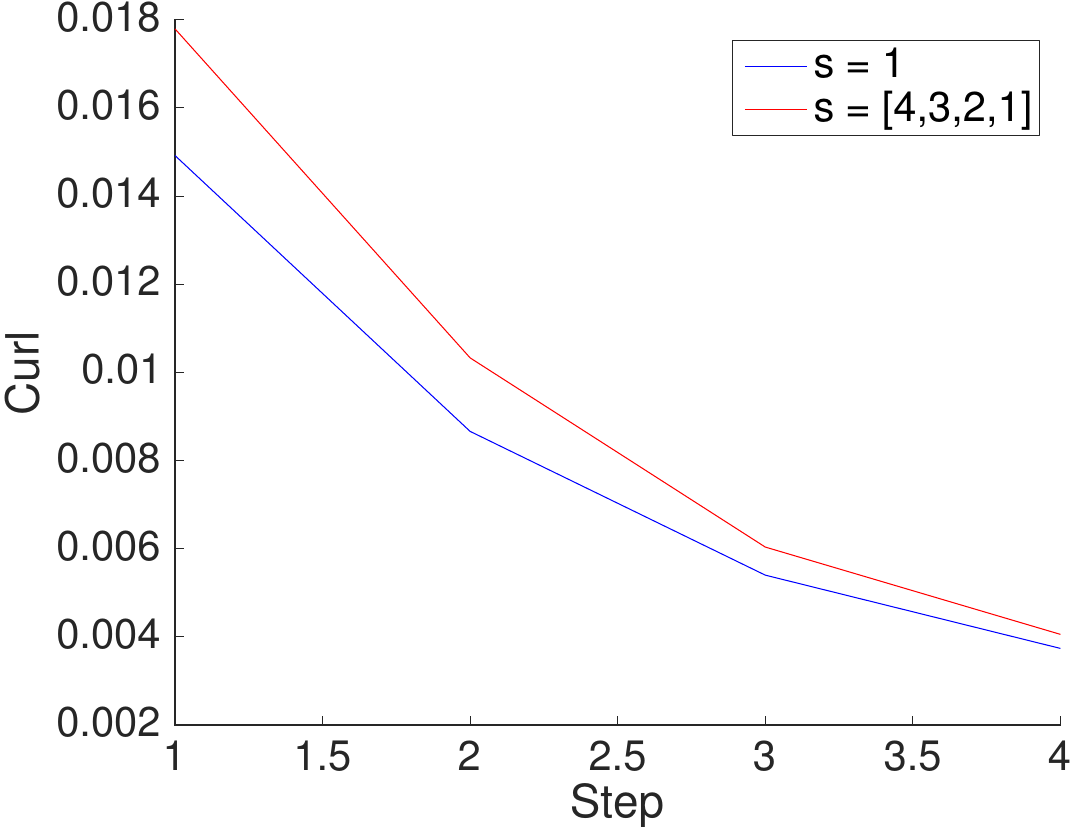}}
  \subfloat[(Absolute) Divergence]{\label{subfig:spatialdiv}\includegraphics[width=0.55\columnwidth]{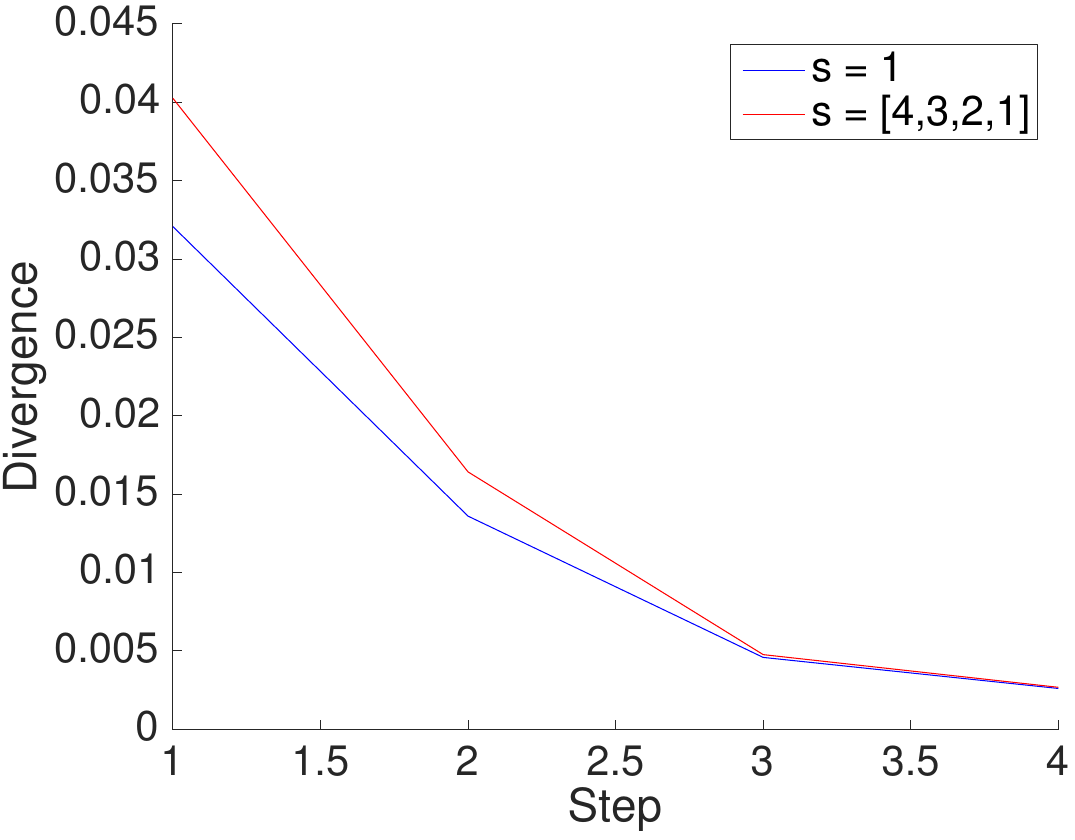}}
  \caption{Results from testing the effect of changing the spatial resolution from low to
    full. The lines are the mean over all points. For the full (blue) and low-to-full
    (red) resolution, the difference in results are (a): blue $1.42\times10^{-4}$ vs red
    $1.58\times10^{-4}$, (b): blue $3.74\times10^{-3}$ vs red $4.21\times10^{-3}$, and
    (c): blue $2.60\times10^{-3}$ vs red $2.70\times10^{-3}$.}
  \label{fig:spatialtest}
\end{figure*}
The spatial resolution is set to $s=[4,3,2,1]$, which is equivalent to smoothly interpolating every 4th point, followed by every 3rd, etc. As with the control point spacing of the deformation field, we scale this to fit the image if the image is not cubical. For instance, if the image has the spatial dimension $100\times150\times50$ then space between spatial interpolations for $s=3$ will be $[2,3,1]$ with a bound on no less than 1 (i.e. full resolution). The results are shown in \Cref{fig:spatialtest}. The results are very similar in the final step. However, the hierarchical resolution approach compared to the full resolution gave a speed-up factor of 2.4.

\subsubsection{A Qualitative Example of the Results}
\label{subsec:qualitativeresult}

\begin{figure*}[!htb]
  \centering \subfloat[Ground
  truth]{\label{subfig:cc_slice_2D}\includegraphics[width=0.6\columnwidth]{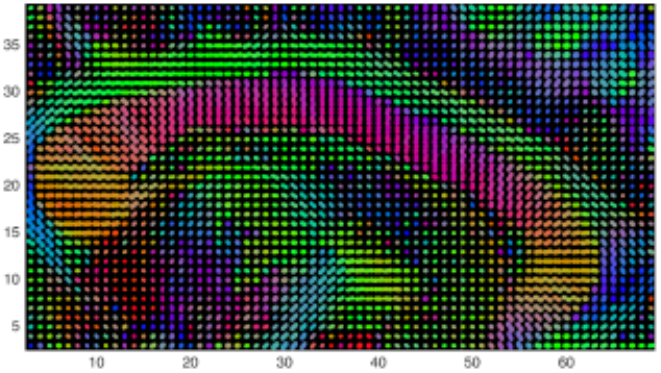}}
  \subfloat[Warped
  image]{\label{subfig:cc_warp_2D}\includegraphics[width=0.6\columnwidth]{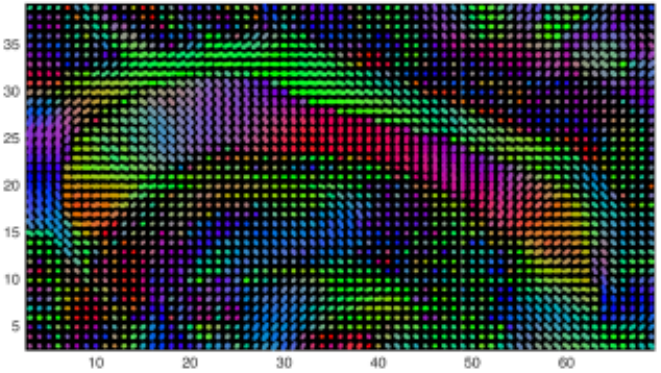}}
  \subfloat[Registered image
  reconstruction]{\label{subfig:cc_warp_2D_registered}\includegraphics[width=0.6\columnwidth]{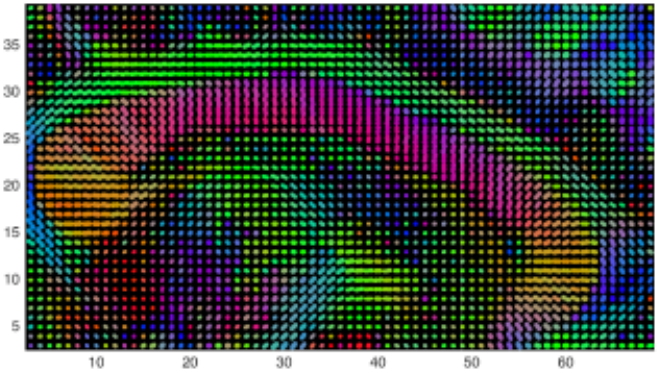}}
  \caption{2D visualization of \Cref{subfig:cc_slice}, and the reconstructed warped image
    after applying the deformation field to the original image. We will be registering (b)
    back to (a).}
  \label{fig:cc_slice_2D_result}
\end{figure*}

\begin{figure}[!t]
  \centering
  \subfloat[Ground truth]{\label{subfig:cc_slice_3D_zoom}\includegraphics[width=0.9\columnwidth]{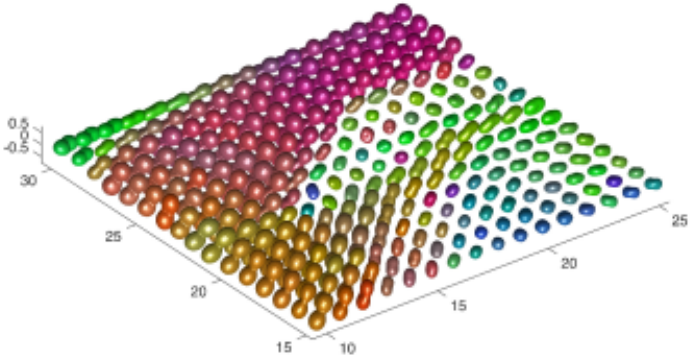}}  \\
  \subfloat[Registered image
  reconstruction]{\label{subfig:cc_warp_3D_zoom_result}\includegraphics[width=0.9\columnwidth]{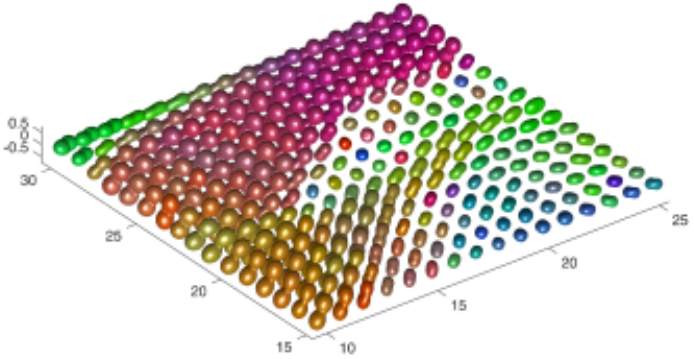}}
  \caption{Same as \Cref{fig:cc_slice_2D_result}, zoomed in on the left side (anterior) of
    the figure (around the Genu). The images are not a 100 percent the same, but the
    difference is hard to notice, and the registration more than adequate.}
  \label{fig:cc_slice_result_zoom}
\end{figure}

Finally, we visualize the registration so as to perform a qualitative assessment of the warp, shape, and orientation of the individual ODFs. All registrations are based on the raw, noise-corrected HARDI data, while we show the tomographic inversion (FRT) indicating the direction of the diffusion and likely fiber tract orientations. Unlike the simulated experiments, we have fitted a B-spline to the image before deforming and visualizing the result, in contrast to the smoothing spline used in the artificial examples. The results where obtained using an increasing number of control point resolution with $\kappa=15$, same bin resolution refining and  spatial resolution refining as above. \Cref{fig:cc_slice_2D_result} shows the result for the central ROI slice, along with a zoomed-in version in \Cref{fig:cc_slice_result_zoom}.

\section{Conclusion}

We have presented a scale-space formulation of density estimation that extends LOR to spatio-directional data, explicitly for the registration of DWI data with an explicit reorientation of the full diffusion profile. We have provided empirical evidence showing that the underlying structure of the data is preserved during registration while providing excellent registration results through many classical artificial examples known to be difficult for registration. We have shown that the formulation of the similarity itself provides regularization through the additional information given by the orientational dimension, illustrated clearly in the artificial examples. We have investigated the different scales provided by the framework and shown how the different parameters influence the registration results. In conclusion, LORD provides a smooth and well-matched deformation of DWI data, and the registration results are improved by the induced regularization obtained by integrating the orientational information in the objective function. Finally, we have shown that the deformation has very little impact on the shape of the deformed ODF.

\section*{Acknowledgements}
This research was supported by Center for Stochastic Geometry and Advanced Bioimaging, funded by grant 8721 from the Villum Foundation.

\bibliographystyle{spmpsci}
\bibliography{IEEEabrv,mybib}
\vspace{-1cm}

\end{document}